\documentclass[10pt, twocolumn, twoside]{IEEEtran}

\usepackage{times}
\usepackage{epsfig}
\usepackage{graphicx}
\usepackage{amsmath}
\usepackage{amssymb}
\usepackage[commentsnumbered]{algorithm2e}
\usepackage{threeparttable}
\usepackage{booktabs, dcolumn}
\usepackage{stfloats}
\usepackage{adjustbox}
\usepackage{array}
\usepackage{booktabs}
\usepackage{multirow}
\usepackage{rotating}
\usepackage{color}
\usepackage{subfig}
\usepackage{tabularx, booktabs}
\newcolumntype{Y}{>{\centering\arraybackslash}X}
\usepackage{float}
\usepackage{pifont}% http://ctan.org/pkg/pifont
\newcommand{\cmark}{\ding{52}}%
\newcommand{\xmark}{\ding{53}}%

\begin{document}

\title{Hierarchical Invariant Feature Learning with Marginalization for Person Re-Identification}

\author{Rahul~Rama~Varior, ~\IEEEmembership{Student Member,~IEEE,}
%        Tian-Tsong~Ng,~\IEEEmembership{Member,~IEEE,}
        ~Gang~Wang,~\IEEEmembership{Member,~IEEE}  % <-this % stops a space
\thanks{R. Rama Varior and G. Wang are with the School of Electrical and Electronic Engineering, Nanyang Technological University, Singapore, 639798. (e-mail: \{rahul004,wanggang\}@ntu.edu.sg.\protect) }}

%\thanks{ G. Wang is also with the Advanced Digital Sciences Center, Singapore, 138632. \protect (e-mail: gang.wang@adsc.com.sg.\protect)}}

%\markboth{Transaction on Image Processing,~Vol.~xx, No.~xx, xx~20xx}%
%{Shell \MakeLowercase{\textit{et al.}}: Bare Demo of IEEEtran.cls
%for Journals}

\maketitle

\begin{abstract}
This paper addresses the problem of matching pedestrians across multiple camera views, known as person re-identification. Variations in lighting conditions, environment and pose changes across camera views make re-identification a challenging problem. Previous methods address these challenges by designing specific features or by learning a distance function. 
%Due to these variations, the ideal metric required for person re-identification becomes highly non-linear. 
We propose a hierarchical feature learning framework that learns invariant representations from labeled image pairs. A mapping is learned such that the extracted features are invariant for images belonging to same individual across views. To learn robust representations and to achieve better generalization to unseen data, the system has to be trained with a large amount of data. Critically, most of the person re-identification datasets are small. Manually augmenting the dataset by partial corruption of input data introduces additional computational burden as it requires several training epochs to converge. We propose a hierarchical network which incorporates a marginalization technique that can reap the benefits of training on large datasets without explicit augmentation. We compare our approach with several baseline algorithms as well as popular linear and non-linear metric learning algorithms and demonstrate improved performance on challenging publicly available datasets, VIPeR, CUHK01, CAVIAR4REID and iLIDS. Our approach also achieves the state-of-the-art results on these datasets.
\end{abstract}

\IEEEpeerreviewmaketitle

\section{Introduction}
\label{sec:intro}
\begin{figure}
\begin{center}
%\fbox{\rule{0pt}{2in} \rule{1\linewidth}{0pt}}
% \includegraphics[width=1\linewidth ] {./fig/exampleimages.pdf}             
 \includegraphics[width=1\linewidth ] {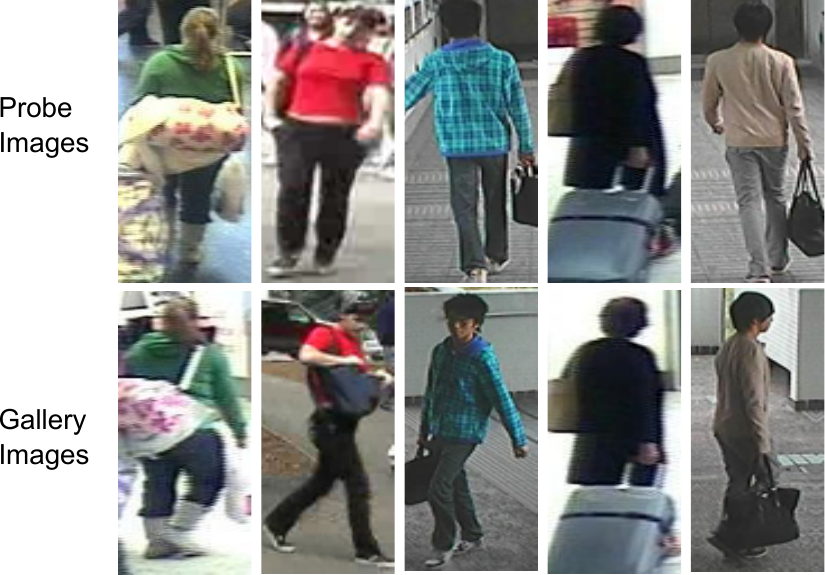}  
\end{center}
   \caption{Some images taken from standard person re-identification datasets such as VIPeR \cite{viper}, CUHK01 \cite{li2012human}, iLIDS \cite{iLIDS} and CAVIAR4REID \cite{caviar}. The objective of person re-identification is to match a given probe image to a set of gallery images. \bf Best viewed in color}
\label{fig:images}
\end{figure}

{M}{atching} pedestrians across multiple non-overlapping camera views is a research problem that has gained a lot of interest in recent years. It has become an integral part of surveillance, human tracking and human retrieval. Figure \ref{fig:images} shows some example images of pedestrians captured from such non-overlapping camera views. The objective of this problem is to identify the matching image(s) from a set of gallery images for a given probe image, thereby saving labor intensive work of searching through the entire set of images for identifying the correct match. Main approaches that address this problem concentrate on developing a feature representation \cite{li2014deepreid,bicovma2012,sdalf} and \cite{salientcolorECCV14} for the images or learning a distance metric \cite{yi2014deep,ZhenliShiyu_CVPR2013,pedagadi2013local} \cite{prdczhengpami2012} and \cite{Weinberger2009LMNN} so that images belonging to the same person are closer to each other in a feature space.

Despite the efforts of several researchers over the years, person re-identification still remains a challenging problem. In this paper, we address two major challenges in person re-identification. First, environmental conditions such as varying illumination, backgrounds and changes in pose across camera views, resolution and image quality cause significant change in appearance for images of same individual across camera views. Previous methods \cite{bicovma2012,viewpointinvariant} address these challenges by designing features specific for each of these aspects. We propose a hierarchical network that can learn invariant representations from labeled image pairs across different views. Local features are first extracted from the images. Inspired by the success of kernel based methods in many computer vision problems \cite{bo2011object} and \cite{bo2010kernel}, we use a non-linear kernel function to map the input features to a kernel space. Further, a linear mapping is learned to extract patterns in this kernel space so that for corresponding parts of matching images, the extracted patterns are {\it close} to each other. To learn the linear mapping, we take advantage of labeled image pairs and enforce the invariance constraint for the mapped kernel features across different views.

The second problem addressed in this paper is the lack of labeled training data. For a good generalization to unseen data, complex systems must be trained with a large amount of data. Most of the existing person re-identification datasets are small. Augmenting the data explicitly by partial corruption based on a specific corruption distribution can be adopted. But augmenting the data makes the system computationally expensive as it requires several epochs over the entire training set to converge. Motivated by the technique used in \cite{chen2014marginalized}, we propose a novel hierarchical network which incorporates the marginalization technique that can reap the benefits of training on augmented dataset without explicitly adding more data. The technique is to marginalize out the noise of the corrupted inputs. The proposed framework is related to the kernel Local Fisher Discriminant Analysis (kLFDA) framework in \cite{eccv14prid} which can be possibly extended to a two layer structure. But a potential difficulty is applying the marginalization technique over the LFDA framework. Therefore, we adapt the SVM Metric Learning (SVMML) in \cite{ZhenliShiyu_CVPR2013} by incorporating the marginalization technique. This leads to an improved metric learning algorithm which suits better for smaller person re-identification datasets.

To the best of our knowledge, this is the first work that proposes a hierarchical invariant feature learning framework coupled with the SVM metric learning framework simultaneously incorporating marginalization technique to address the problem of lack of training data for person re-identification.
In short, main contributions of this paper are; 
\begin{itemize}
  \item We propose a new learning framework that captures invariant information from local patches of image pairs for person re-identification. A linear transformation is learned in the kernel space by enforcing invariance constraint for local patch features extracted from labeled image pairs. %The proposed method employs multiple non-linear transformations by transforming the input features into a kernel space with the extracted local features as anchor points. Further, these exemplar features in the kernel space undergo a linear transformation so that invariant information is captured. 
 
  \item We propose a novel feature learning framework that incorporates marginalization that can reap the benefits of training on {\it infinite} data without explicitly augmenting them. At the second layer of the proposed hierarchical network, we adapt the SVMML by incorporating marginalization. Marginalization helps in achieving better generalization over unseen data.
  
  \item We show that the proposed method outperforms several baselines and popular kernel based algorithms for VIPeR \cite{viper}, CUHK01\cite{li2012human}, CAVIAR4REID \cite{caviar} and iLIDS \cite{iLIDS} re-identification datasets. Our approach also achieves state-of-the-art results on these datasets.
\end{itemize}

The rest of the paper is organized as follows. Section \ref{sec:rel} gives a brief overview of the related works on metric learning and feature learning for person re-identification as well as some of the recent deep learning frameworks. Detailed description of our approach with formulations are given in Section \ref{framework}. In Section \ref{sec:exp}, we show our experimental results. We analyze different baselines in Section \ref{sec:analysis} and Section \ref{sec:concl} concludes this paper.

\section{Related Work}
\label{sec:rel}
Existing works for person re-identification focus on several aspects of the problem such as developing a feature representation \cite{wang2014person,bicovma2012,sdalf,salientcolorECCV14,yangcolor2014}, metric learning  \cite{ZhenliShiyu_CVPR2013,yi2014deep,prdczhengpami2012,pedagadi2013local} for distance computation and learning mid-level representations \cite{layne2012towards,li2014deepreid,zhao2014learning}. \cite{zhao2013person,zhao2013unsupervised} focus on finding the salient patches and rank the matching images using rank SVM. 
%In another recent work, person re-identification is treated as a bipartite graph matching problem \cite{zhang2014structured}. 
Majority of the works \cite{bicovma2012,viewpointinvariant,zhao2013unsupervised} focus on developing feature representation based on texture, color, shape, regions and interest points. Below we give some of the related works in Metric Learning and Feature Learning for person re-identification

\subsection{Metric Learning}
Some of the prominent metric learning algorithms proposed for person re-identification are \cite{li2012human,kissmecvpr12,ZhenliShiyu_CVPR2013,pedagadi2013local,prdczhengpami2012} etc.
Regularized LFDA was proposed for person re-identification in \cite{pedagadi2013local}. The objective is to maximize the inter-class separability and to minimize the within class variance. To address the non-linearities in feature space, it was proposed to use kernel based dimensionality reduction techniques in \cite{pridwithlearnedmetric}. SVM Metric learning was proposed in \cite{ZhenliShiyu_CVPR2013}, and the idea is to learn a decision boundary that is adaptive to the data samples. In \cite{eccv14prid}, several kernel-based metric learning methods for person re-identification were evaluated and kernel based LFDA was found to be performing the best for several re-identification datasets. Deep metric learning architectures has also been proposed for person re-identification in \cite{yi2014deep} and \cite{li2014deepreid}. In \cite{yi2014deep}, data augmentation is employed for improved performance. Deep architectures have to be trained with a large amount of data for better generalization to unseen data. However, none of the above metric learning algorithms address the problem of lack of labeled training data in re-identification datasets. Hence, we propose a marginalized metric learning framework based on SVMML objective function instead of techniques such as artificial augmentation of the dataset. 

\subsection{Feature Learning}

In \cite{li2014deepreid} a deep architecture was proposed to learn filter pairs that can handle photometric and geometric transformations. It consists of a single convolutional layer with  max-pooling and a patch matching layer that multiplies the feature responses of the convolutional layer at different horizontal offsets. \cite{yi2014deep} also contains a feature learning framework based on Convolutional Neural Networks (CNN). The similarity of the features from the CNN is measured using a Cosine similarity function. A recent work \cite{ejazdeep2015} also proposes a CNN based deep architecture for person re-identification. In addition to convolution and max-pooling, a Cross-Input Neighborhood Difference is employed to compute relationships between images from two views. In another related work \cite{zhao2014learning}, mid-level filters were learned from patch clusters to achieve cross view invariance. But variations in illumination, viewing angle and other environmental variables affect the cross view invariance of the features as well as demand non-linear mappings for the features. Therefore, we propose an invariant feature learning framework in kernel space which can be seen as learning a flexible non-linear transformation in the original feature space \cite{wang2014two}. This can be very effective for overcoming the non-linearities due to the aforementioned challenges. In addition to that, deep architectures required enormous amount of data to learn robust representations. Our approach also has the advantage of training on large amount of data without explicit augmentation as we employ marginalization.

%\subsection{Kernel methods}
%Kernel based methods have achieved great success in computer vision problems to generate image level representations which can be used in conjunction with linear classifiers to achieve state-of-the-art in several object recognition datasets \cite{bo2011object,bo2010kernel}. Kernel descriptors are applied over kernel descriptors in a hierarchical manner to form image-level representations in \cite{bo2011object}. Several kernels were also evaluated in \cite{eccv14prid}. Inspired by the success of these ideas, in our work, we map the local features into a kernel space by using an advanced kernel function with the local features as anchor points. However, different from these works, we take advantage of image pairs to learn invariant features which is more suitable for person re-identification. Moreover, the hierarchical model proposed in this paper can reap the benefits of large training datasets.
% Further, a linear transformation is applied over the features in the kernel space to capture invariant information from image pairs. 

\subsection{Deep Learning}

Some of the prominent articles in deep learning \cite{bengio2009learning,vincent2010stacked,krizhevsky2012imagenet} suggest that multiple layers with non-linear transformations can be efficient in directly modeling complex functions mapping the input to output. Several successful algorithms have been proposed \cite{hinton2006fast,hinton2006reducing,bengio2007greedy,bengio2009learning} to train such large networks such as deep belief networks and stacked auto-encoders. In-order to achieve a better generalization, such systems need to be trained with a large amount of data. Artificial augmentation of the data by partial corruption was proposed in \cite{vincent2008extracting}. But augmented data creates additional computational burden. To circumvent the complexity due to augmented data, marginalized Denoising Auto-Encoder ({\bf mDAE})  was proposed in \cite{chen2014marginalized,chen2012marginalized}. This is a variant of the traditional denoising auto-encoder \cite{vincent2010stacked}. The key technique is to marginalize out the noise by adding a regularization term and thereby reaping the benefits of training on {\it infinite} data without explicitly corrupting the original data. Inspired by this work, we propose a novel marginalized invariant feature learning framework as well as marginalized metric learning framework based on the \textbf{SVMML} formulation. 
\section{Approach}
\label{framework}

\begin{figure*}
\begin{center}
%\fbox{\rule{0pt}{2in} \rule{1\linewidth}{0pt}}
 \includegraphics[width=1\linewidth ]
                   {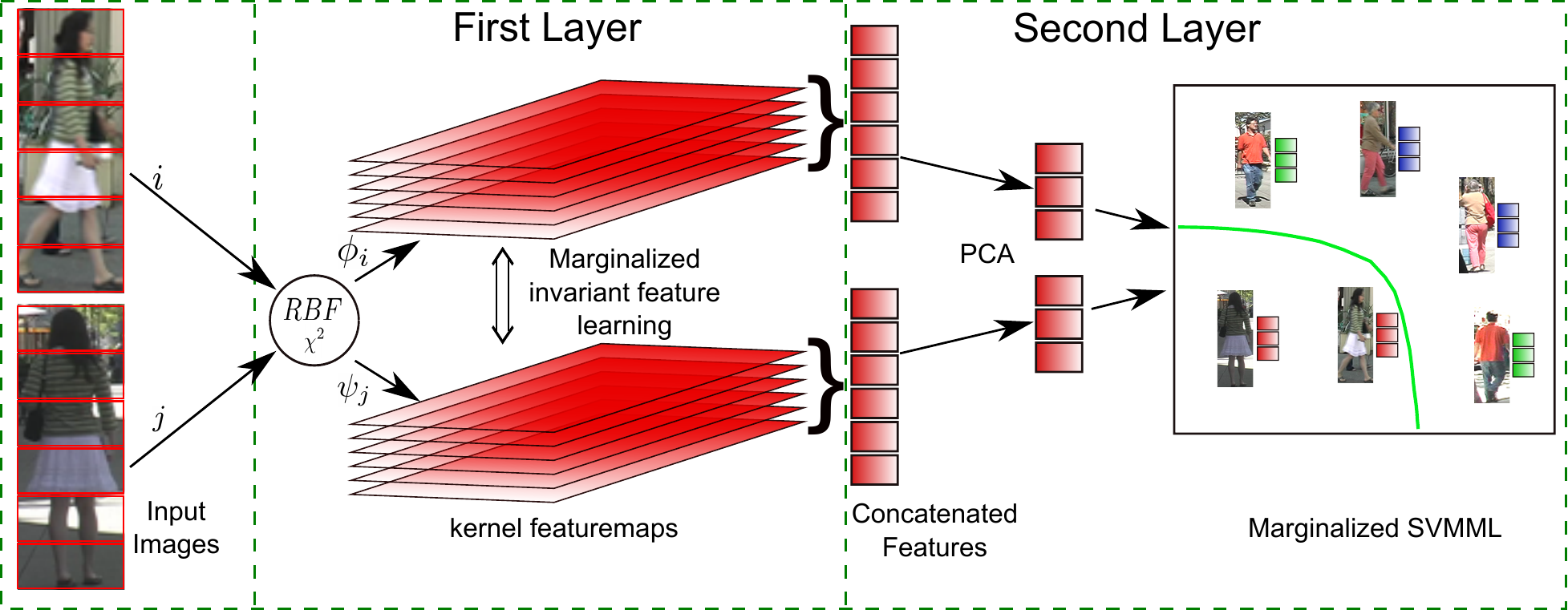}             
%                   {fig/framework/framework.pdf}  
\end{center}
   \caption{The process pipeline of the proposed framework. Images are divided into horizontal stripes and features are extracted. By using the $D$ exemplars the RBF-$\chi^2$ kernel mapping is performed. Further these features are fed into a marginalized invariant feature learning framework to extract invariant features. After the feature extraction, the representation of individual stripes are concatenated to form the whole image representation and a PCA is used to reduce the dimensionality of the image representation. Further, marginalized SVMML formulation is used to learn the metric and to rank the matches. \bf Best viewed in color}
\label{fig:framework}
\end{figure*}

%Several multi-layer architectures have been proposed in computer vision. Such multi-layer architectures can extract non-trivial feature maps or representations from the input images and encode complex image characteristics. As mentioned in section \ref{sec:intro}, the 
The proposed framework has a two-layer structure. In the first layer, the input features are first mapped into a kernel space. A linear mapping is learned to extract stable structures and patterns from the exemplar responses. Due to environmental conditions such as varying illumination, backgrounds and changes in pose across camera views, successful re-identification requires invariant features from different views. Therefore, the extracted patterns must be {\it close} to each other for corresponding parts of matching image pairs. This is achieved by enforcing an invariance constraint to the learned features. We do not extract any discriminative information by using negative pairs at this stage as the discriminative capability of local stripes may be insufficient. %As mentioned in \cite{yi2014deep}, variations in pose, illumination and other environmental and camera parameters make the ideal metric for person re-identification highly non-linear. Further, \cite{wang2014two} suggests that learning a linear mapping in a kernel space can be treated as learning a flexible non-linear mapping in the original space. Therefore, the input features are transformed into a kernel space and a linear mapping is learned in the new space which is equivalent to a non-linear transformation of the input features. 
Additionally, since most of the person re-identification datasets are small, labeled data is scarce. Therefore, the overall objective function is also adapted by the marginalization technique which further boosts the generalization capability. Further, the image representation is obtained by concatenating the mapped local features of that image. In the second layer, the features of the whole image is fed into a metric learning framework. We adapt the SVMML framework by incorporating marginalization so that with the limited amount of training data available, better performance can be achieved. The visualization of the process pipeline is shown in figure \ref{fig:framework}. Below, we explain our approach in detail.

\subsection{Features}
\label{features}
Each image is divided equally into $6$ non-overlapping horizontal stripes as in \cite{pcca,prdczhengpami2012} and \cite{eccv14prid}. For each of these horizontal stripes, following features were extracted. 
\subsubsection{LBP}
Texture patterns were captured by Local Binary Pattern (LBP) \cite{lbp} histograms computed with 8-neighbors at a radius 1 and 16-neighbors at a radius 2. The dimensions of the representations are $59$ and $243$ for 8-neighbors and 16-neighbors respectively.
\subsubsection{Color Histograms}
Color information was captured by computing 16 bin histograms for each of the RGB, HS and YUV color channels respectively. The total dimension of the color representation will be 128.
 
An $l1$ normalization is performed for each of these channels individually and concatenated to form the final feature representation. This leads to a $430$ dimensional representation for each of the horizontal stripes. Computing the histogram over horizontal stripes can give the advantage of translational invariance which is significant in person re-identification problems due to the pose changes. For a fair comparison, these features were used for all the performance comparison against the baseline methods as well as the metric learning methods.
%\subsection{Kernel mapping based on exemplars}
%\label{kmboe}
Once the local features (horizontal stripes) are obtained, these features are projected into a kernel space by using them as anchor points (exemplars) and a linear transformation is learned by enforcing an invariance constraint.

\subsection{Learning invariant features}

%As mentioned in Section \ref{kmboe}, learning a linear mapping in the kernel space is equivalent to learning a flexible non-linear transformation in the original space. After computing the kernel responses, a linear transformation is learned from them in the kernel space 
The objective of learning a linear transformation in the kernel space is to capture stable structures and patterns which are invariant across views.
%Here, {\it important} patterns mean the stable structures in the form of dependencies and regularities characteristic of the unknown distribution of its observed input. 
%Such patterns can be captured from the inputs by using an auto-encoder which learns a mapping or a set of filters from the reconstruction error. We adapt the traditional auto-encoder formulation so that the captured information is also invariant for corresponding parts of the matching images. 
Let $z_{\phi_i}$ and $z_{\psi_j}$ be the mapped responses for the corresponding stripes of two matching images in the probe and gallery respectively. They can be mathematically expressed as 
%\begin{eqnarray}
%z_i = (W^{(1)} \phi_i + b^{(1)} ) 
%\end{eqnarray} 
%\begin{eqnarray}
%z_j = (W^{(3)} \psi_j + b^{(3)} ) 
%\end{eqnarray} 
\begin{eqnarray}
\label{map1}
z_{\phi_i} = f_\theta(\phi_i) = W^{(1)} \phi_i + b^{(1)}
\end{eqnarray} 
\begin{eqnarray}
\label{map2}
z_{\psi_j} = f_\pi(\psi_j) = W^{(3)} \psi_j + b^{(3)}
\end{eqnarray}

where $\phi_i\in\mathbb{R}^{D}$ denotes the exemplar response vector of one of the stripes of a probe image and $\psi_j\in\mathbb{R}^{D}$ denotes the exemplar response vector of the corresponding matching stripe of the matching image pair in the gallery. $W^{(1)}$, $b^{(1)}$, $W^{(3)}$ and $b^{(3)}$ are the learned linear transformation parameters. Since the probe and gallery images are from different sources, we learn separate transformations for each of them as in \cite{li2014deepreid,liu2014semi}. The objective of the mapping is to ensure $z_{\phi_i}$ and $z_{\psi_j}$ to be {\it close} to each other. Thus the overall loss can be formulated as

\begin{eqnarray}
\label{eqnbasic}
\nonumber l (\phi_i,f_\theta(\phi_i), \psi_j,f_\pi(\psi_j)) & = & \left \| \phi_i - (W^{(2)} z_{\phi_i} + b^{(2)} )  \right \|^2\\ \nonumber &+& \left \| \psi_j - (W^{(4)} z_{\psi_j} + b^{(4)} )  \right \|^2\\ \nonumber &+& 
\left \| z_{\phi_i} - z_{\psi_j} \right \|^2_2 \\ 
%\nonumber &+& \lambda \left( \left \|W^{(3)}\right \|_2^2 + \left \| W^{(4)} \right \|^2_2 \right)
%\\  &+& \lambda \left( \left \|W^{(1)}\right \|_2^2 +  \left \| W^{(2)} \right \|^2_2 \right)
\end{eqnarray} 

Here, $W^{(1)}, b^{(1)}, W^{(2)}$ and $b^{(2)}$ are the weight and bias parameters for the transformation and reconstruction steps for probe exemplar response vector and $W^{(3)}, b^{(3)}, W^{(4)}$ and $b^{(4)}$ are the weight and bias parameters for the transformation and reconstruction steps for gallery exemplar response vector.
The first and second terms on the right hand side (RHS) of (\ref{eqnbasic}) are the reconstruction terms for the inputs. These terms are essential to avoid learning trivial solutions for the weight and bias parameters. $f_\theta(\phi_i)$ and $f_\pi(\psi_j)$ denotes the mappings shown in equation (\ref{map1}) and (\ref{map2}) respectively. The objective function was split for the probe and gallery and an alternative optimization scheme was adopted since the invariance term contains parameters from two sources. The final objective functions to be solved for the probe and gallery training data can be written as

%\begin{eqnarray}
%l(\phi_i,f_\theta(\phi_i), \psi_j) &=& \sum_{d}^{ } \left \{ \phi^d - (w_{d}^{(2)} z + b_d^{(2)} )  \right \}^2\\ \nonumber &+& 
%\left \|\sum_{d}^{ } (w_{d}^{(1)} \phi_i^d) - \sum_{d}^{ } (w_{d}^{(1)} \psi_j^d)  \right \|^2_2
%\end{eqnarray} 
%\begin{eqnarray}
%\label{eqnprob}
%l (\phi_i,f_\theta(\phi_i), \psi_j) & = & \alpha \left \| \phi_i - (W^{(2)} z_i + b^{(2)} )  \right \|^2\\ \nonumber &+& 
%\beta \left \| z_i - z_j \right \|^2_2 \\ \nonumber &+& \lambda \left \|W^{(1)}\right \|_2^2 + \lambda \left \| W^{(2)} \right \|^2_2
%\end{eqnarray} 
%
%\begin{eqnarray}
%\label{eqngallery}
%l (\psi_j,f_\theta(\psi_j), \phi_i) & = & \alpha \left \| \psi_j - (W^{(4)} z_j + b^{(4)} )  \right \|^2\\ \nonumber &+& 
%\beta \left \| z_i - z_j \right \|^2_2 \\ \nonumber &+& \lambda \left \|W^{(3)}\right \|_2^2 + \lambda \left \| W^{(4)} \right \|^2_2
%\end{eqnarray}  

\begin{eqnarray}
\label{eqnprob}
\nonumber l (\phi_i,f_\theta(\phi_i)) & = & \left \| \phi_i - (W^{(2)} z_{\phi_i} + b^{(2)} )  \right \|^2\\  &+& 
\left \| z_{\phi_i} - z_{\psi_j} \right \|^2_2 
%\\  &+& \lambda \left \|W^{(1)}\right \|_2^2 + \lambda \left \| W^{(2)} \right \|^2_2
\end{eqnarray} 

\begin{eqnarray}
\label{eqngallery}
\nonumber l (\psi_i,f_\pi(\psi_i)) & = &  \left \| \psi_i - (W^{(4)} z_{\psi_i} + b^{(4)} )  \right \|^2\\  &+& 
\left \| z_{\phi_i} - z_{\psi_j} \right \|^2_2 
%\\  &+& \lambda \left \|W^{(3)}\right \|_2^2 + \lambda \left \| W^{(4)} \right \|^2_2
\end{eqnarray} 

%where $\phi_i$ and $\psi_i$ are the probe and gallery features respectively in the kernel space. $z_{\phi_i}$ and $z_{\psi_i}$ are the corresponding mapped responses.

\subsection{Marginalization}

To learn a robust representation and to have good generalization capability, the system needs to be trained with a large amount of data. Since most of the person re-identification datasets are small, the learned transformations will most likely overfit and may not have good generalization capability to unseen data. The concept of denoising auto-encoder ({\bf DAE}) \cite{vincent2008extracting} was introduced to learn robust representation from limited training data. In a DAE, the training data is artificially corrupted based on a specific corruption distribution to create more training data. The corrupted data is sampled from the conditional distribution $p(\tilde{\phi}|\phi )$ and the commonly used corruption distribution is the additive Gaussian distribution. Let $\tilde{\phi_i}$ be the corrupted copy of $\phi_i$. 
%$\tilde{\phi_i}$ is initially mapped to the latent representation by using the encoder. Further it is mapped to the original space using the decoder and the error with respect to the original data $\phi_i$ is computed. The objective is to minimize this error so that robust representations can be obtained even from corrupted data and the resulting learned transformation will have better generalization capabilities. But 
The process of explicitly corrupting the data $\phi_i$ creates multiple copies $\tilde{\phi_{i}^1}, ..., \tilde{\phi_{i}^m}$ of the data and the total amount of data becomes $m$ fold. For a large $m$, the computational complexity increases to a large extent as it requires several epochs over the entire training set to converge. %To handle the $m$-fold dataset without corrupting the inputs explicitly is challenging. 

Marginalized Denoising Auto-Encoder ({\bf mDAE}) \cite{chen2014marginalized} was proposed to address the additional computational complexity due to augmented data. The technique is to marginalize out the corruption during auto-encoder training. In the proposed framework, the key difference for our objective function is that, in addition to the auto-encoder loss, an invariance constraint is enforced between the learned representations to achieve cross-view invariance. Therefore, we adapt the mDAE by incorporating the invariance term in our framework. Besides, the proposed framework consists of two parallel networks paired by the invariance term. Therefore it requires an alternate optimization scheme. Below, we present the derivation for one of the parallel networks and for the other network, it can be worked out similarly.
%In this case, the average loss over the entire corpus along with the corrupted data becomes the expectation of the loss as ${m\to\infty}$. 

As ${m\to\infty}$, the average loss over the entire corpus along with the corrupted data becomes the expectation of the loss. Therefore optimizing the objective for {\it infinite} training data is equivalent to optimizing the below objective function.

\begin{eqnarray}
\frac{1}{n}\sum_{i=1}^{n}\mathrm{E}_{p(\tilde{\phi_i}|\phi_i)}\left [ l(\phi_i,f_\theta(\tilde \phi_i)) \right ]
\end{eqnarray}

where $\tilde \phi_i$ is the corrupted version of the $i^{th}$ exemplar. $l(.)$ denotes the proposed loss function. 
%For the proposed approach, since the loss function is adapted with the invariance term, the average loss becomes
%\begin{eqnarray}
%\frac{1}{n}\sum_{i=1}^{n}\mathrm{E}_{p(\tilde{\phi_i}|\phi_i)}\left [ l(\phi_i,f_\theta(\tilde \phi_i), \psi_j) \right ]
%\end{eqnarray}
The loss function is approximated by its Taylor expansion at the mean of the corruption, $\mu_\phi = \mathrm{E}_{p(\tilde{\phi}|\phi)}[\tilde{\phi}]$. 
The proposed loss function is a sum of the Auto-Encoder loss and Invariance loss. The Taylor expansion of each of these losses can be considered independently and the following derivation can be applied to both of them.

For simplicity of notation, here $\phi\in\mathbb{R}^D$ denotes one of the data points in a training set of $\Phi = \{\phi_1,...,\phi_n\}$ and $\tilde{\phi}$ denotes its corrupted version. The Taylor expansion yields the following learning objective.

\begin{eqnarray}
\label{taylorobj}
 \nonumber l(\phi,f_\theta(\tilde \phi)) &\approx & l(\phi,f_\theta(\mu_{\phi})) \\ \nonumber 
&+& (\tilde \phi -\mu_{\phi})^T\bigtriangledown_{\tilde \phi} l\\
&+& \frac{1}{2} (\tilde \phi - \mu_{\phi})^T \bigtriangledown_{\tilde \phi} ^2 l (\tilde \phi - \mu_{\phi}) 
\end{eqnarray}
Taking the expectation on both sides for (\ref{taylorobj}),

\begin{eqnarray}
\label{expectationlhsrhs}
\nonumber \mathrm{E}_{p(\tilde{\phi}|\phi)}\left [ l(\phi,f_\theta(\tilde \phi)) \right ] \approx  l(\phi,f_\theta(\mu_{\phi})) \\\nonumber + \mathrm{E}_{p(\tilde{\phi}|\phi)}\left [(\tilde \phi - \mu_{\phi})^T \right ] \bigtriangledown_{\tilde \phi} l \\ + \frac{1}{2}  tr \left (\mathrm{E}\left [ (\tilde \phi - \mu_{\phi})(\tilde \phi - \mu_{\phi})^T \right ]   \bigtriangledown_{\tilde \phi} ^2 l \right )
\end{eqnarray}

The second term on the RHS of (\ref{expectationlhsrhs}) vanishes since the expectation of the left hand side (LHS) yields $ \mathrm{E}_{p(\tilde{\phi}|\phi)}[\tilde{\phi}] = \mu_{\phi}$. Therefore, Taylor expansion at the mean $\mu_\phi$ is critical. Further simplification results in the following objective.

\begin{eqnarray}
\label{exp}
 \mathrm{E}\left [ l(\phi,f_\theta(\tilde \phi)) \right ]\approx l(\phi,f_\theta( \mu_\phi)) + \frac{1}{2} tr \left ( \Sigma_{\phi} \bigtriangledown_{\tilde \phi} ^2 l \right )
\end{eqnarray}

where $\Sigma_\phi = \mathrm{E}\left [ (\tilde \phi - \mu_{\phi})(\tilde \phi - \mu_{\phi})^T \right ]$ is the variance of the corruption distribution and $ \bigtriangledown_{\tilde \phi} ^2 l $ is the Hessian of $l(.)$ with respect to $\tilde \phi$. $\Sigma_\phi$ is a diagonal matrix as the assumption is that the corruption is applied to each dimension independently. Therefore, computing the Hessian for higher dimensional data simplifies to computing only its diagonal elements. The $d^{th}$ dimension of the Hessian matrix's diagonal can be obtained by the straightforward application of chain rule as shown below.

\begin{eqnarray}
\label{chainrule}
\nonumber\frac{\partial^2 l}{\partial (\tilde{\phi}(d))^2} &=& \left ( \frac{\partial z}{\partial \tilde \phi(d)} \right )^T\frac{\partial^2 l}{\partial z^2}\left ( \frac{\partial z}{\partial \tilde \phi(d)} \right )\\ &+&\left ( \frac{\partial l}{\partial z}\right )^T\frac{\partial^2 z}{\partial   (\tilde\phi(d))^2}
\end{eqnarray}

where $\tilde{\phi}(d)$ represents the $d^{th}$ dimension of $\tilde{\phi}$ and $z$ is the latent representation. The corruption is applied to each dimension of $\phi$ independently, hence the second derivative with respect to each dimension of $\phi$.

Following \cite{lecun98b}, the last term in (\ref{chainrule}) can be dropped and the Hessian can be approximated as 

\begin{eqnarray}
\label{marginalization}
\frac{\partial^2 l}{\partial (\tilde{\phi}(d))^2} \approx \sum_{h=1}^{D_h}\frac{\partial^2 l}{\partial z_h^2}\left ( \frac{\partial z_h}{\partial \tilde \phi(d)} \right )^2
\end{eqnarray}

Substituting (\ref{marginalization}) in (\ref{exp}) yields the objective function below.

\begin{eqnarray}
\label{margDAE}
l(\phi,f_\theta(\mu_{\phi})) + \frac{1}{2} \sum_{d=1}^{D}\sigma ^2_{\phi(d)}\sum_{h=1}^{D_h}\frac{\partial^2 l}{\partial z_h^2}\left ( \frac{\partial z_h}{\partial \tilde \phi(d)} \right )^2
\end{eqnarray}
where $D$ and $D_h$ are the input and hidden layer dimensions respectively. $z_h$ denotes the $h^{th}$ dimension of the latent representation of the input. The value of $\sigma_{\phi(d)}^2$ for additive Gaussian is $\sigma^2_d $ and the mean $\mu_{\phi}$ is $\phi$.

Further, we need to compute the second term on the RHS of (\ref{margDAE}) for the proposed loss function. Since the proposed loss function consists of the Auto-Encoder and Invariance terms, it involves computing $\frac{\partial^2 l}{\partial (\tilde{\phi}(d))^2}$ for both the Auto-encoder term as well as the Invariance term. Details are given below. 

The final objective function is obtained by adapting the objective function in (\ref{eqnprob}) and (\ref{eqngallery}) according to the marginalization technique. 
%As explained above, the second term on the RHS of equation (\ref{taylorobj}) vanishes in the Taylor expansion as the expectation of the loss function is taken. Rest of the derivation until the approximation of the chain rule in equation (\ref{marginalization}) remains same. 
Let $\mathrm{R}_\theta^p$ be the marginalization term obtained by applying (\ref{marginalization}) over (\ref{eqnprob}) for the probe data and $\mathrm{R}_\pi^g$ be the corresponding marginalization term for the gallery data. The exemplar responses $\phi$ and $\psi$ respectively for the probe and gallery input features are fed into two separate networks paired by the invariance term. The objective function for the probe ($l^{(p)}$) can be obtained by substituting $\mathrm{R}_\theta^p$ in (\ref{margDAE}). 
\vspace{-1pt}
\begin{eqnarray}
\label{eqn9}
\nonumber l^{(p)} &=& l(\phi,f_\theta({{\phi}})) + \frac{1}{2}\sum_{d=1}^{D}\sigma_d^2 \mathrm{R}_\theta^p \\ \nonumber
&=& l(\phi,f_\theta({{\phi}})) +  \sum_{d=1}^{D}\sigma_d^2\left (\sum_{h}^{ } (w_{hd}^{(1)})^2  \right )  \nonumber \\ &+&
\sum_{d=1}^{D}\sigma_d^2 \sum_{h}^{} \left (\left (\sum_{d}^{ } (w_{hd}^{(2)})^2  \right )(w_{hd}^{(1)})^2  \right )
\end{eqnarray}
Similarly, the objective function for gallery ($l^{(g)}$) is as shown below.
\vspace{-1pt}
\begin{eqnarray}
\label{eqn10}
\nonumber l^{(g)} &=& l(\psi,f_\pi({{\psi}})) + \frac{1}{2}\sum_{d=1}^{D}\sigma_d^2 \mathrm{R}_\pi^g \\ \nonumber
&=& l(\psi,f_\pi({{\psi}})) + \sum_{d=1}^{D}\sigma_d^2\left (\sum_{h}^{ } (w_{hd}^{(3)})^2  \right )  \nonumber \\ & +& \sum_{d=1}^{D}\sigma_d^2 \sum_{h}^{} \left (\left (\sum_{d}^{ } (w_{hd}^{(4)})^2  \right )(w_{hd}^{(3)})^2   \right )
\end{eqnarray}

where $w_{hd}^{(1)}$ corresponds to an element in the matrix $W^{(1)}$ and similarly for the other matrices.

The last term in (\ref{eqn9}) and (\ref{eqn10}) can be obtained from the marginalized denoising auto-encoder cost function $l(\phi,f_\theta(\tilde \phi))$ as shown in \cite{chen2014marginalized}. The marginalization penalty for the invariance term can be derived by taking the second derivative of the invariance loss function with respect to the input $\phi$ for each of its dimensions $d$. Let $l_{inv}$ be the invariance term.  The derivation with respect to a data point is shown below.
%$z_{\phi_i}$ and $z_{\psi_j}$

\begin{multline}
  l_{inv} = \left \| z_{\phi_i} - z_{\psi_j} \right \|^2_2 \\  =   \sum_{h}^{ }\left (\sum_{d}^{ } (w_{hd}^{(1)} \phi_i(d)) + b^{(1)}_h - \sum_{d}^{ } (w_{hd}^{(3)} \psi_j(d)) + b^{(3)}_h  \right )^2 
\end{multline} 

\vspace{-10pt}
\begin{multline}
 \frac{\partial l_{inv}}{\partial \tilde{\phi_i}(d)} =   \\  2 \sum_{h}^{ }\left (\sum_{d}^{ } (w_{hd}^{(1)} \phi_i(d)) + b^{(1)}_h - \sum_{d}^{ } (w_{hd}^{(3)} \psi_j(d)) + b^{(3)}_h  \right )w_{hd}^{(1)} \\
\end{multline}

 \begin{eqnarray}
 \label{invmarg}
 \frac{\partial^2 l_{inv}}{\partial (\tilde{\phi_i}(d))^2}= 2 \left (\sum_{h}^{ } (w_{hd}^{(1)})^2  \right ) 
 \end{eqnarray}

The derivation of $\left(\frac{\partial^2 l_{inv}}{\partial (\tilde{\psi_i}(d))^2} \right)$ can be done as shown above which will result in the second term on the RHS of (\ref{eqn10}).
%\vspace{-10pt}
\begin{eqnarray}
\label{marginv2}
 \frac{\partial^2 l_{inv}}{\partial (\tilde{\psi}(d))^2}= 2 \left (\sum_{h}^{ } (w_{hd}^{(3)})^2  \right ) 
\end{eqnarray}

Further, the weight decay term, $\lambda \left \|W^{(i)} \right \|^2_2$ for $\{i = 1, \cdots, 4\}$ is also added to the respective objective functions, (\ref{eqn9}) and (\ref{eqn10}) with a penalty $\lambda$. Hence the final objective function can be given as 
\newline

For Probe

\begin{eqnarray}
\label{eqn9fin}
 \nonumber l^{(p)} & = & \left \| \phi_i - (W^{(2)} z_{\phi_i} + b^{(2)} )  \right \|^2 + 
 \left \| z_{\phi_i} - z_{\psi_j} \right \|^2_2 \\ \nonumber &+&   \sum_{d=1}^{D}\sigma_d^2\left (\sum_{h}^{ } (w_{hd}^{(1)})^2  \right ) \\ \nonumber &+&
 \sum_{d=1}^{D}\sigma_d^2  \left (\sum_{d}^{ } (w_{d}^{(2)})^2  \right )(w_{hd}^{(1)})^2  \\  &+&  \lambda \left \|W^{(1)}\right \|^2_2 + \lambda \left \| W^{(2)} \right \|^2_2 
\end{eqnarray}

For Gallery 
%\vspace{-4pt}
\begin{eqnarray}
\label{eqn10fin}
\nonumber   l^{(g)} & =&   \left \| \psi_i - (W^{(4)} z_{\psi_i} + b^{(4)} )  \right \|^2+
\left \| z_{\phi_i} - z_{\psi_j} \right \|^2_2  \\ \nonumber &+&  \sum_{d=1}^{D}\sigma_d^2\left (\sum_{h}^{ } (w_{hd}^{(3)})^2  \right )  \\ \nonumber  &+& \sum_{d=1}^{D}\sigma_d^2 \left (\sum_{d}^{ } (w_{d}^{(4)})^2  \right )(w_{hd}^{(3)})^2  \\ &+&    \lambda \left \|W^{(3)}\right \|^2_2 + \lambda \left \|W^{(4)} \right \|^2_2 
\end{eqnarray}
%
%\vspace{-1pt}
% \begin{eqnarray}
% \label{marginv1}
% & & \frac{\partial^2 l_{inv}}{\partial (\tilde{\phi}(d))^2}= 2 \left (\sum_{h}^{ } (w_{hd}^{(1)})^2  \right ) 
% \end{eqnarray}
% \vspace{-5pt}
% \begin{eqnarray}
% \label{marginv2}
% & & \frac{\partial^2 l_{inv}}{\partial (\tilde{\psi}(d))^2}= 2 \left (\sum_{h}^{ } (w_{hd}^{(3)})^2  \right ) 
% \end{eqnarray}
%Detailed derivation for (\ref{marginv1}) and (\ref{marginv2}) is given in the supplementary material. 
%Further, the weight decay term, $\lambda \left \|W^{(i)} \right \|^2_2$ for $\{i = 1, \cdots, 4\}$ is also added to the respective objective functions, (\ref{eqn9}) and (\ref{eqn10}) with a penalty $\lambda$.
%\subsection{Second layer}
%
%The features obtained from the first layer, $z_\phi$ and $z_\psi$ are used to form the representation of the probe and gallery images respectively. The mapped exemplar responses of the corresponding $6$ local regions are concatenated to form the representation for the whole image. These image representations are the inputs to the second layer. In the second layer, we perform the non-linear RBF-$\chi^2$ transformation again. Finally, for the mapped features from the second layer, the within class variance has to be reduced and the between class variance has to be maximized. This is achieved by using Local Fisher Discriminant Analysis framework. Other metric learning algorithms can also be used in this layer, but we chose LFDA which proves to be the best among different metric learning algorithms for a given set of features as shown in \cite{eccv14prid}.

\subsection{Metric Learning}

The second layer of the proposed system is a marginalized Metric Learning framework based on the SVM Metric Learning (SVMML). The objective of SVMML is to compute a decision boundary which is locally adaptive to the data samples. In our framework, we adapt the traditional SVMML by incorporating marginalization so that the benefits of training on large amount of data can be achieved. The objective function in equations (\ref{eqn9fin}) and (\ref{eqn10fin}) are solved alternatively and the parameters for the probe and gallery are learned. By using equations (\ref{map1}) and (\ref{map2}), the probe and gallery exemplars are mapped into an invariant feature space. Once $z_{\phi_i}$ and $z_{\psi_i}$ are obtained, the global image representation is obtained by concatenating the mapped exemplar responses of the $6$ horizontal stripes in the image. The global image representation is projected into a lower dimensional space by PCA. Let $D_{ml}$ be the dimension of the projected space. These act as inputs to the proposed framework.

Let $k_i\in\mathbb{R}^{D_{ml}}$ be the global image representation of a probe image and $k_{i'}\in\mathbb{R}^{D_{ml}}$ be an image representation from the gallery set. The second order decision function as given in \cite{ZhenliShiyu_CVPR2013} can be written as 
\begin{eqnarray}
\label{mlloss}
\nonumber f(k_i, k_{i'}) &=& \frac{1}{2}k_{i}^T A k_{i} + \frac{1}{2}k_{i'}^T A k_{i'} + k_{i}^T B k_{i'} \\ &+&  c^T(k_{i} + k_{i'}) + b
\end{eqnarray}

where $A$ and $B$ are real, symmetric, positive semi definite (PSD) and negative semi definite (NSD) matrices respectively. $c$ is a $d$ - dimensional vector and $b$ is the bias term. In practice, the authors of \cite{ZhenliShiyu_CVPR2013} apply a $log - exp$ transformation to the loss function in equation (\ref{mlloss}) and also omit the term $c^T(k_{i} + k_{i'})$ in their implementation.

Therefore, the final objective function of SVMML is

\begin{eqnarray}
\label{ml}
l_{ml} = g(f(k_i, k_{i'})) = log(e^{f(k_i, k_{i'})} + 1) 
\end{eqnarray}
where $g(.)$ denotes the $log - exp$ transformation.

Eventhough $k_i$ and $k_{i'}$ are mapped responses from two sources, the invariance term projects them into the same feature space. Hence the marginalization can be directly applied to the loss function in equation (\ref{ml}) without having two separate objective functions for probe and gallery data.

Similar to the derivation of the marginalization term for the invariance loss in equation (\ref{invmarg}), the second derivative of $l_{ml}$ with respect to each dimension of $k_i$ has to be computed. Below, we show the derivation w.r.t one training example. This can be generalized to the entire training dataset and the final loss function for the metric learning framework can be obtained.

\begin{eqnarray}
\frac{\partial l_{ml}}{\partial \tilde{k_i}(d)} =    \frac{1}{2}\frac{\partial g}{\partial {f}} \mathbin{{\odot}} (Ak_i + A^T k_i + k_{i'}^T B^T)
\end{eqnarray}

Here, $\odot$ denotes the element-wise multiplication. The second derivative of $l_{ml}$ w.r.t each dimension of $k_i$ is given by,
%\\ \nonumber &&

 \begin{multline}
 \label{mlmarg}
  \frac{\partial^2 l_{ml}}{\partial (\tilde{k_i}(d))^2}= \frac{1}{2}\frac{\partial g}{\partial {f}} \mathbin{{\odot}} csum(A + A^T)   + \\  \frac{1}{2}\frac{\partial^2 g}{\partial {f^2}} \mathbin{{\odot}} (Ak_i + A^T k_i + k_{i'}^T B^T) \mathbin{{\odot}}(Ak_i + A^T k_i + k_{i'}^T B^T) 
 \end{multline}

Here, $csum(.)$ denotes the column-wise sum of a matrix. The partial derivatives 

 \begin{eqnarray}
 \label{pdv}
 \nonumber \frac{\partial g}{\partial {f}} = \frac{1}{1 + e^{-f(k_i, k_{i'})}}
 \end{eqnarray}

 \begin{eqnarray}
 \label{pdv2}
 \nonumber \frac{\partial^2 g}{\partial {f^2}}  =  \left(\frac{1}{1 + e^{-f(k_i, k_{i'})}} \right)\times \left( \frac{1}{1 + e^{f(k_i, k_{i'})}} \right)
 \end{eqnarray}
 
The final loss function for the metric learning framework can be obtained by substituting equation (\ref{mlmarg}) in equation (\ref{ml})
%\begin{eqnarray}
%\label{mlfin}
%\nonumber l_{ml} &=& log(e^{f(k_i, k_{i'})} + 1) + \frac{1}{2} \sum_{d=1}^{D_{ml}}\frac{\partial^2 l_{ml}}{\partial (\tilde{k_i}(d))^2} \\ \nonumber &=& log(e^{f(k_i, k_{i'})} + 1) + \frac{1}{2}\frac{\partial g}{\partial {f}} \mathbin{{\odot}} csum(A + A^T)    \\ \nonumber &+ &\frac{1}{2}\frac{\partial^2 g}{\partial {f^2}} \mathbin{{\odot}} (Ak_i + A^T k_i + k_{i'}^T B^T) \\ && \hspace{70pt} \mathbin{{\odot}}(Ak_i + A^T k_i + k_{i'}^T B^T)
%\end{eqnarray}
\begin{eqnarray}
\label{mlfin}
 l_{ml} = log(e^{f(k_i, k_{i'})} + 1) + \frac{1}{2} \sum_{d=1}^{D_{ml}}\sigma ^2_{k_i}\frac{\partial^2 l_{ml}}{\partial (\tilde{k_i}(d))^2} 
\end{eqnarray}
$D_{ml}$ denotes the dimension of the image representation. We do not use a low-rank projection for the metric learning framework. However, decomposing A and B to $A = M M^T$ and $B = - N N^T$ can be helpful in learning a PSD and NSD low-rank matrices respectively. Further Frobenius norm regularization for $A$ and $B$ were added to the objective function in equation (\ref{mlfin}).
\subsection{Optimization}
The objective functions in (\ref{eqn9}) and (\ref{eqn10}) are minimized alternatively for the parameters $W^{(i)}$ and $b^{(i)}$ in (\ref{eqnprob}) and (\ref{eqngallery}) respectively. The first network, as explained in (\ref{eqn9}) is optimized for $\kappa$ iterations while keeping the parameters of (\ref{eqn10}) fixed and vice-versa. L-BFGS gradient based minimization is adopted for optimizing the cost function and the total number of iterations is kept as $300$. The main parameters of the experiment were empirically determined by cross-validation as done in \cite{eccv14prid} on the VIPeR dataset and kept same for other datasets. They are, the dimension of the linearly projected space $D_h = 800$, $\lambda = 1 \times 10^{-7}$, $\sigma_d = 0.1$. 

Objective function in equation (\ref{mlfin}) can be optimized by gradient projection algorithms. If a low-rank projection is required, optimization has to be done for $M$ and $N$. Similar to the first layer, number of iterations is kept as $300$.	Main parameters are determined by cross-validation on the VIPeR dataset and kept same for others. They are, dimension of the global image representation $D_{ml} = 400$, $\sigma_{k_i}=0.01$ and the frobenius regularization penalty , $\lambda_{ml} = \{1\times10^{-8}, 1\times10^{-7}\}$ for $A$ and $B$ respectively.

%\subsubsection{LFDA}

\section{Experiments}
\label{sec:exp}

Our approach was evaluated on four challenging publicly available datasets which are characterized by ample variation in their environments, pose and illumination. For a fair comparison, we use the same features and experimental settings for all the baselines. The amount of supervision for each of the datasets is also kept the same and the results are reported as Cumulative Matching Characteristics (CMC), which shows the probability of identifying the correct match at different ranks. 
%We also compare our approach with some of the baselines which can be considered as variants of the proposed approach as well as other popular kernel based metric learning algorithms. 

%\subsubsection{Comparisons}
The main baselines which can be considered as the variants of our approach are listed below. %\newline \newline

\begin{enumerate}
%\item {\bf lin\_Kernel} - We experiment using the same architecture with a linear kernel to prove that non-linear kernel transformations are required since variations in illumination, pose background and other camera parameters are prominent. At the second layer, we use SVMML.

\item {\bf 2 layer kLFDA} - To show that the proposed architecture with marginalization technique has better generalization capability over unseen data, we develop a baseline by extending the kLFDA \cite{eccv14prid} into two layers.

\item {\bf no\_Marg} - To prove that marginalization technique helps in achieving better performance, we compare the proposed approach to its variant without marginalization at both layers. We use SVMML at the second layer for a fair comparison. This makes the overall a system an autoencoder with invariance term coupled with the traditional SVMML metric learning framework

\item {\bf no\_Inv} - To show that the proposed objective function with invariance is crucial for person re-identification performance, we conduct experiments without enforcing the invariance constraint in our objective function. Removing the invariance term makes the first layer objective function a simple marginalized Denoising Auto-encoder. At the second layer, the proposed marginalized SVMML is used to learn the metric.

\item {\bf marg\_SVMML} - To prove that the proposed marginalized metric learning framework achieves better performance, we couple the proposed marginalized framework in the first layer with the traditional SVMML.

\item {\bf marg\_kLFDA} - This framework is similar to the above. Instead of using the traditional SVMML framework at the second layer, we use kLFDA at the second layer.

%\item {\bf Ours with single layer} - To show that the hierarchical structure with multiple non-linear transformation is better than single layer network, we compare the proposed approach with its variant by removing the second layer's RBF-$\chi^2$ transformation and directly feeding the image representation to the classifier.
\end{enumerate}
\begin{table*}[h!t]

	\renewcommand{\arraystretch}{1.3}
	\setlength{\tabcolsep}{2.5pt}
	\caption{Summary of the baseline approaches.\\ L1 - Layer - I , L2 - Layer II , Kernel - \xmark \hspace{4pt} means No kernel is used}
	\label{table_base}
	\centering
	{
	\begin{tabular}{|c|c|c|c|c|c|c|c|}
		\hline
		\bfseries Baseline & \bfseries Kernel - L1 &  \bfseries Invariance - L1 & \bfseries Marginalization - L1 & \bfseries Kernel - L2 & \bfseries Marginalization - L2 & \bfseries Metric Learning
		\\\hline\hline
%		{\begin{tabular}{@{}c@{}}lin\_Kernel\end{tabular}} 		&  Linear  		 &  \cmark & \cmark  &  \xmark  &  \cmark  &  SVMML \\
%		\hline		
		{\begin{tabular}{@{}c@{}}2 layer kLFDA\end{tabular}} 	&  RBF-$\chi^2$  &  \xmark & \xmark  &   RBF-$\chi^2$  &  \xmark  &  kLFDA  \\
		\hline
		{\begin{tabular}{@{}c@{}}no\_Marg\end{tabular}}  		&  RBF-$\chi^2$  &  \cmark & \xmark  &  \xmark  &  \xmark  &  SVMML  \\
		\hline
		{\begin{tabular}{@{}c@{}}no\_Inv\end{tabular}} 			&  RBF-$\chi^2$  &  \xmark & \cmark  &  \xmark  &  \cmark  &  SVMML  \\
		\hline
		{ \begin{tabular}{@{}c@{}} marg\_SVMML\end{tabular}} 	&  RBF-$\chi^2$  &  \cmark & \cmark  &  \xmark  &  \xmark  &  SVMML  \\
		\hline		
		{ \begin{tabular}{@{}c@{}} marg\_kLFDA\end{tabular}} 	&  RBF-$\chi^2$  &  \cmark & \cmark  &  RBF-$\chi^2$  &  \xmark  &  kLFDA  \\
		\hline
		{\begin{tabular}{@{}c@{}} Ours \end{tabular}} 			&  RBF-$\chi^2$  &  \cmark & \cmark  &  \xmark  &  \cmark  &  SVMML  \\
		\hline	
			
	\end{tabular}	}
\end{table*}
A summary of the above baselines is given in table \ref{table_base} for easy reference.
In addition to the above baselines, our approach was also compared with popular linear and kernel based metric learning algorithms. Finally, we compare the proposed algorithm with the state-of-the-art algorithms in all datasets.

\subsection{Evaluation Methodology}

All the experiments were conducted in the single-shot setting as done in \cite{eccv14prid,ensemble2015}, i.e. for each query image from the probe set is compared to one image from the gallery set. For all the datasets, the images are divided into $6$ non-overlapping horizontal stripes and the features are extracted as explained in section \ref{features}. Since the datasets are small, all the horizontal stripes from the training images are used as anchor points (exemplars) for the RBF-$\chi^2$ kernel mapping. Further, the a transformation is learned by solving the objective functions in equations (\ref{eqn9}) and (\ref{eqn10}) with the invariance constraint between the exemplar pairs in the kernel space. The linear transformation projects the mapped kernel responses to an $800$ dimensional space. To obtain the global image representation, the mapped responses of the $6$ horizontal stripes are concatenated. PCA is employed to reduce the $4800$ dimensional responses to a $400$ dimensional representation. Finally, the global image representation is fed into the marginalized metric learning framework to learn the metric. For all the comparison with the baselines as well as popular metric learning algorithms, the features as mentioned in section \ref{features} were used. 

\subsection{Datasets}

\subsubsection{VIPeR Dataset}

VIPeR dataset \cite{viper} is one of the most challenging datasets for person re-identification. It contains 1264 images of 632 pedestrians captured from two different camera views. Image resolution is 128$\times$48 and significant variations in illumination and pose can be observed in this dataset. In our experiments and for the comparisons in the tables, $316$ image pairs were used as training images. %The evaluation settings are kept the same as in \cite{eccv14prid}.
For each of the training image pairs, feature pairs are generated which results in a total of 1896 such pairs. All the local features were used as exemplars and features were transformed into the kernel space which led to a $3792$ dimensional representation for each of the stripes. 

We compare our approach with popular kernel based and other non-linear metric learning algorithms proposed for person re-identification and it can be seen from Table \ref{viper_base} that our method outperforms all the other metric learning approaches for person re-identification. Table \ref{viper_actbase} shows the performance comparison of our approach against the baselines. It can be seen that the proposed invariant feature learning framework with marginalization can achieve better results than its variants. 

 The performance comparison with the state-of-the-art methods is shown in Table \ref{viper_state}. The proposed algorithm beats all the other methods individually at all ranks. Combining different methods with complementary features and ensemble of metric learning algorithms have also been studied before. As shown in table \ref{viper_state}, the metric ensembles proposed in \cite{ensemble2015} gives a matching rate of $45.9\%$ at Rank 1. Combination of our approach with \cite{ZhenliShiyu_CVPR2013} outperforms all the existing methods for VIPeR dataset achieving state-of-the-art results. Results show that our features are complementary to the features used in \cite{ZhenliShiyu_CVPR2013}. It should be noted that our approach alone has comparable performance with the results obtained by \cite{ensemble2015} and the combination of \cite{zhao2014learning} and \cite{ZhenliShiyu_CVPR2013} at higher ranks.

\begin{table*}[!ht]

	\renewcommand{\arraystretch}{1.2}
	
	\caption{Performance Comparison of our approach with the baseline algorithms on the VIPeR, CAVIAR4REID, iLIDS and CUHK01 datasets. Proposed framework outperform all the baselines and other variants of this method except for VIPeR dataset where marg\_kLFDA performs slightly better than our method at rank 20.}
	\centering
		\label{actual_baseline}
	\subfloat[ViPER]{
		\begin{tabular}{|c|c|c|c|c|c|}
%		\label{viper_Base}
			\hline
			\bfseries Method & \bfseries Rank 1 & \bfseries Rank 5 & \bfseries Rank 10 & \bfseries Rank 20
			\\\hline\hline
%			{\begin{tabular}{@{}c@{}}lin\_Kernel\end{tabular}}  & 23.7  &  57.9  &  72.2  &  83.5 \\
%			\hline		
			{\begin{tabular}{@{}c@{}}2 layer kLFDA\end{tabular}}  & 32.7  & 65.2  &  79.1  &  90.2 \\
			\hline
			{\begin{tabular}{@{}c@{}}no\_Inv\end{tabular}}  &  35.3  & 70.6  & 82.5  & 91.7 \\
			\hline
			{\begin{tabular}{@{}c@{}}no\_Marg\end{tabular}}  & 33.2 &  67.7 &  79.8    &  90.9 \\
			\hline
			{ \begin{tabular}{@{}c@{}} marg\_SVMML\end{tabular}} & 34.8 &  69.4 &  82.6 &  91.2 \\
			\hline		
			{ \begin{tabular}{@{}c@{}} marg\_kLFDA\end{tabular}} & { 36.4}   & { 70.4}  & { 83.6}  & { \bf 93.4} \\
			\hline
			{\bf \begin{tabular}{@{}c@{}} Ours \end{tabular}} & { \bf39.3}   & {\bf 73.0}  & { \bf84.6}  & { 92.5} \\
			\hline	
%			{\bf  \begin{tabular}{@{}c@{}} Ours* \end{tabular}} & { \bf 41.5}   & { \bf 75.3}  & { \bf 85.8}  & { 93.0} \\
%			\hline				
		\end{tabular}
		\label{viper_actbase}
	}\quad
	\subfloat[CAVIAR4REID]{
	\begin{tabular}{|c|c|c|c|c|c|}
		\hline
		\bfseries Method & \bfseries Rank 1 & \bfseries Rank 5 & \bfseries Rank 10 & \bfseries Rank 20
		\\\hline\hline
%		{\begin{tabular}{@{}c@{}}lin\_Kernel\end{tabular}}  & 30.7  &  62.5  &  79.0  &  92.7\\
%		\hline		
		{\begin{tabular}{@{}c@{}}2 layer kLFDA\end{tabular}}  & 36.7   &    65.2     &   77.8   &  92.1 \\
		\hline	
		{\begin{tabular}{@{}c@{}}no\_Inv\end{tabular}}  & 36.8 &  70.8 &  85.7  & 95.7 \\
		\hline				
		{\begin{tabular}{@{}c@{}}no\_Marg\end{tabular}}  & 34.1 &    70.0  &  83.9  &  93.9 \\
		\hline
		{\begin{tabular}{@{}c@{}} marg\_SVMML\end{tabular}} & 38.3 &  71.3  & 84.6  & 95.2		 \\
		\hline		
		{\begin{tabular}{@{}c@{}} marg\_kLFDA\end{tabular}} & { \bf 40.2 }  & 73.1  &  85.9  &  {\bf96.3} \\
		\hline
		{\bf \begin{tabular}{@{}c@{}} Ours \end{tabular}} & {  39.9}   & { \bf 73.0}  & { \bf 86.9}  & {  95.7} \\
		\hline	
%		{\bf \begin{tabular}{@{}c@{}} Ours* \end{tabular}} & 40.1  & { \bf 73.6}  & { \bf 88.4}  & { \bf 96.3} \\
%		\hline					
	\end{tabular}
	\label{caviar_actbase}
	}	
	
	\subfloat[iLIDS]{
		\begin{tabular}{|c|c|c|c|c|c|}
			\hline
			\bfseries Method & \bfseries Rank 1 & \bfseries Rank 5 & \bfseries Rank 10 & \bfseries Rank 20
			\\\hline\hline		
%			{\begin{tabular}{@{}c@{}}lin\_Kernel\end{tabular}}  & 30.7    & 60.9   &  70.4   &  86.0 \\
%			\hline		
			{\begin{tabular}{@{}c@{}}2 layer kLFDA\end{tabular}}  & 37.3  & 65.1  &  76.5  &  89.3 \\
			\hline	
			{\begin{tabular}{@{}c@{}}no\_Inv\end{tabular}}  & 37.0    & 65.9  &  77.8  &  88.3 \\
			\hline			
			{\begin{tabular}{@{}c@{}}no\_Marg\end{tabular}}  & 35.0  & 62.5  & 75.6  &  87.5 \\
			\hline
			{ \begin{tabular}{@{}c@{}} marg\_SVMML\end{tabular}} & 36.1  &  66.3  &  78.9  &  89.5 \\
			\hline		
			{ \begin{tabular}{@{}c@{}} marg\_kLFDA\end{tabular}} & 39.1  &  68.4  & {\bf 81.7 } &  91.2 \\
			\hline		
			{\bf \begin{tabular}{@{}c@{}} Ours\end{tabular}} & { \bf  39.5}   & { \bf 70.4}  & {  81.0}  & { \bf 91.4} \\
			\hline
%			{\bf \begin{tabular}{@{}c@{}} Ours*\end{tabular}} & { \bf  40.7}   & { \bf 72.2}  & {  81.5}  & { \bf 92.0} \\
%			\hline			
		\end{tabular}
		\label{ilids_actbase}
	}	\quad
	\subfloat[CUHK01]{
	\begin{tabular}{|c|c|c|c|c|c|}
		\hline
		\bfseries Method & \bfseries Rank 1 & \bfseries Rank 5 & \bfseries Rank 10 & \bfseries Rank 20
		\\\hline\hline		
%		{\begin{tabular}{@{}c@{}}lin\_Kernel\end{tabular}}  & 17.3 &  39.4 &  48.9 &  61.2 \\
%		\hline		
		{\begin{tabular}{@{}c@{}}2 layer kLFDA\end{tabular}}  & 26.8 &  49.6  & 60.4 &  71.3 \\
		\hline	
		{\begin{tabular}{@{}c@{}}no\_Inv\end{tabular}}  & 26.5  & 52.9  & 64.0  & 74.5 \\
		\hline			
		{\begin{tabular}{@{}c@{}}no\_Marg\end{tabular}}  & 25.6 &  48.9  & 59.0  & 70.1 \\
		\hline
		{\begin{tabular}{@{}c@{}}marg\_SVMML\end{tabular}}  &  27.2 &  49.4 &  62.1 &  71.8 \\
		\hline
		{\begin{tabular}{@{}c@{}}marg\_kLFDA\end{tabular}}  & 28.2 &  50.5  & 60.8 &  71.5 \\
		\hline				
		{\bf \begin{tabular}{@{}c@{}} Ours\end{tabular}} & { \bf29.2} & { \bf 54.7 }& { \bf 66.3 }  &{ \bf 77.6} \\
%		{\bf \begin{tabular}{@{}c@{}} Proposed \\ approach\end{tabular}} & 39.1  &  68.4  &  81.7  &  91.2 \\
		\hline
%		{\bf \begin{tabular}{@{}c@{}} Ours*\end{tabular}} & { \bf  29.3} &   { \bf 55.9}  & { \bf  66.9}  & { \bf  77.8} \\
%		\hline		
	\end{tabular}
	\label{cuhk_actbase}
	}	
\end{table*}

\begin{table*}[!ht]

	\renewcommand{\arraystretch}{1.2}
	
	\caption{Performance Comparison of our approach with popular linear and kernel based metric learning algorithms on the VIPeR, CAVIAR4REID, iLIDS and CUHK01 datasets. Proposed framework outperform all the other algorithms except for CAVIAR4REID dataset where rPCCA performs slightly better than our method at rank 20.}
	\label{table_Base}
	\centering
	\subfloat[ViPER]{
	\begin{tabular}{|c|c|c|c|c|c|}
		\hline
		\bfseries Method & \bfseries Rank 1 & \bfseries Rank 5 & \bfseries Rank 10 & \bfseries Rank 20
		\\\hline\hline
		PCCA \cite{pcca}  & 19.6  & 51.5  & 68.2  & 82.9  \\
		\hline
		rPCCA  & 22.0  & 54.8  & 71.0  & 85.3  \\
		\hline
		{\begin{tabular}{@{}c@{}}LFDA \\ w/o kernel\end{tabular}}  & 19.7  & 46.7  & 62.1  &  77.0 \\
		\hline
		KISSME \cite{kissmecvpr12}  & 23.8  & 52.9  & 67.1  & 80.5 \\
		\hline
		SVMML & 27.0  & 60.9  & 75.4  & 87.3 \\
		\hline
		kLFDA & 32.3  & 65.8  & 79.7  & 90.9 \\
		\hline
		MFA & 32.2  & 66.0  & 79.7  & 90.6 \\
		\hline
		{\bf \begin{tabular}{@{}c@{}} Ours \end{tabular}} & { \bf 39.3}   & { \bf 73.0}  & {\bf  84.6}  & { \bf 92.5} \\
		\hline
%		{\bf \begin{tabular}{@{}c@{}} Ours* \end{tabular}} & { \bf 41.5}   & { \bf 75.3}  & { \bf 85.8}  & { \bf 93.0} \\
%		\hline		
	\end{tabular}
	\label{viper_base}
	} \quad
	\subfloat[CAVIAR4REID]{
	\begin{tabular}{|c|c|c|c|c|c|}
		\hline
		\bfseries Method & \bfseries Rank 1 & \bfseries Rank 5 & \bfseries Rank 10 & \bfseries Rank 20
		\\\hline\hline
		PCCA \cite{pcca}  & 33.4  & 67.2  & 83.1  & 95.7  \\
		\hline
		rPCCA  & 34.0  & 67.5  & 83.4  &{ \bf 95.8 } \\
		\hline
		{\begin{tabular}{@{}c@{}}LFDA \\ w/o kernel\end{tabular}}  & 31.7  & 56.1  & 70.4  &  86.9 \\
		\hline
		KISSME \cite{kissmecvpr12}  & 31.4  & 61.9  & 77.8  & 92.5 \\
		\hline
		SVMML & 25.8  & 61.4  & 78.6  & 93.6 \\
		\hline
		kLFDA & 35.9  & 63.6  & 77.9  & 91.2 \\
		\hline
		MFA & 38.4  & 69.0  & 83.6  & 95.1 \\
		\hline
		{\bf \begin{tabular}{@{}c@{}} Ours \end{tabular}} & { \bf 39.9}   & { \bf 73.0}  & { \bf 86.9}  & {95.7} \\
		\hline		
	\end{tabular}
	\label{caviar_base}
	}	
	
	\subfloat[iLIDS]{
	\begin{tabular}{|c|c|c|c|c|c|}
		\hline
		\bfseries Method & \bfseries Rank 1 & \bfseries Rank 5 & \bfseries Rank 10 & \bfseries Rank 20
		\\\hline\hline
		PCCA \cite{pcca}  & 24.1  & 53.3  & 69.2  & 84.8  \\
		\hline
		rPCCA  & 28.0  & 56.5  & 71.8  & 85.9  \\
		\hline
		{\begin{tabular}{@{}c@{}}LFDA \\ w/o kernel\end{tabular}}  & 32.2  & 56.0  & 68.7  &  81.6 \\
		\hline
		KISSME \cite{kissmecvpr12}  & 28.0  & 54.2  & 67.9  & 81.6 \\
		\hline
		SVMML & 20.8  & 49.1  & 65.4  & 81.7 \\
		\hline
		kLFDA & 36.9  & 65.3  & 78.3  & 89.4 \\
		\hline
		MFA & 32.1  & 58.8  & 72.2  & 85.9 \\
		\hline	
		{\bf \begin{tabular}{@{}c@{}} Ours\end{tabular}} & { \bf  39.5}   & { \bf 70.4}  & { \bf 81.0}  & { \bf 90.7} \\
		\hline				
	\end{tabular}
	\label{ilids_base}
	}\quad	
	\subfloat[CUHK01]{
	\begin{tabular}{|c|c|c|c|c|c|}
		\hline
		\bfseries Method & \bfseries Rank 1 & \bfseries Rank 5 & \bfseries Rank 10 & \bfseries Rank 20
		\\\hline\hline
		PCCA \cite{pcca}  & 17.9 &  41.2 &  54.8  & 69.3  \\
		\hline
		rPCCA  & 21.8 &  47.9&   60.8  & 73.8  \\
		\hline
		{\begin{tabular}{@{}c@{}}LFDA \\ w/o kernel\end{tabular}}  & 15.7  & 34.4 &  44.6  & 56.6 \\
		\hline
		KISSME \cite{kissmecvpr12}  & 10.3 &  27.2 &  37.5  & 49.7 \\
		\hline
		SVMML & 13.5 &  32.5  & 43.7 &  57.3 \\
		\hline
%		kLFDA & {32.9}   & {  57.6}  & {  68.5}  & {  79.7}\\
		kLFDA	&	26.1 &  49.4 &  58.4  & 71.8 \\
		\hline
%		MFA & 30.2 &  56.1  & 67.4  & 77.8		 \\
		MFA	&	27.2  & 47.7 &  58.4  & 70.2\\
		\hline	
		{\bf \begin{tabular}{@{}c@{}} Ours\end{tabular}} & { \bf29.2 }& { \bf 54.7 }& { \bf 66.3}   & { \bf77.6} \\
		\hline			
	\end{tabular}
	\label{cuhk_base}
	}	
\end{table*}

\subsubsection{CAVIAR4REID Dataset}

CAVIAR4REID \cite{caviar} is another challenging dataset for evaluating person re-identification algorithms. This dataset is extracted from the well known CAVIAR dataset for evaluating pedestrian tracking and detection algorithms. It contains a total of 1220 images of 72 different individuals captured from arbitrary viewpoints under varying illumination. Among the 72 pedestrians, 50 of them appear in both camera views and the remaining 22 appear only in one camera view. The image resolution varies from 17 $\times$ 39 to 72 $\times$ 144. In our experiments, we resize each image into 128 $\times$ 48 and extract the features as mentioned in Section \ref{features}. For all the comparisons, the dataset was split into halves for training and testing which results in images of $36$ pedestrians for training and single-shot experiment setting was adopted \cite{eccv14prid}.

%Table \ref{caviar_base} shows the comparison of our approach with the different kernel based and other non-linear metric learning algorithms and it can be seen that our approach outperforms all of them. 
Performance comparison of our approach against the baselines is shown in table \ref{caviar_actbase}. Proposed  algorithm outperforms all the baselines. The performance comparison against popular linear and non-linear metric learning algorithms are shown in table \ref{caviar_base}. We also compare our approach to the state-of-the-art approaches for single-shot setting and show the results in Table \ref{caviar_state}. It can be seen that MFA achieves slightly better result than ours at rank 1. However, at higher ranks we outperform all the other methods. The proposed algorithm combined with the kLFDA \cite{eccv14prid} outperforms all the other methods at all ranks and sets the state-of-the-art performance in the single shot setting for this dataset.

\subsubsection{iLIDS Dataset}

i-LIDS is another multi-shot re-identification dataset captured at a busy airport arrival hall. There are a total of $476$ images of $119$ people. The number of images per person varies from 2 to 8. The images undergo large illumination change, considerable change in view angle and are largely occluded which makes this dataset more challenging. For all the comparisons, the dataset was split into halves for training and testing which results in images of $59$ pedestrians for training. 

Table \ref{ilids_actbase} shows the performance comparison of our method against the baseline approaches. It can be seen that the proposed learning framework achieves the best results. Table \ref{ilids_base} shows the performance comparison of the proposed algorithm with other metric learning algorithms. Table \ref{ilids_state} shows the comparison of our approach with the state-of-the-art approaches in single-shot setting for this dataset. The proposed algorithm individually is comparable to \cite{ensemble2015} at higher ranks. Combination of our algorithm with kLFDA in \cite{eccv14prid} achieves state-of-the-art results at higher ranks for this dataset, but at rank 1, we observed that \cite{ensemble2015} outperforms ours. We believe that this is due to an efficient mechanism to learn weights for individual features or learned metric for boosting rank 1 performance, proposed in \cite{ensemble2015}.

\subsubsection{CUHK01 Dataset}

CUHK01 is a re-identification dataset with $3884$ images of $971$ individuals captured at two different views in a campus environment. For each individual, there are two images in each view. All the images are manually cropped and normalized to $160\times60$ pixels. Variation in pose, illumination makes this re-identification dataset considerably challenging. The dataset is split into halves for training and testing which leads to $485$ individuals for training and rest for testing. 

Table \ref{cuhk_actbase} shows the performance comparison of our method against the baseline approaches. It can be seen that the proposed learning framework achieves the best results. Table \ref{cuhk_base} shows the performance comparison of our method with other metric learning algorithms. Table \ref{cuhk_state} shows the comparison of our approach with the state-of-the-art approaches in single-shot setting. It can be seen that some recent works \cite{ensemble2015} outperforms a combination of ours and \cite{eccv14prid} for iLIDS dataset at Rank 1 and CUHK01 dataset at all ranks. 
\begin{table*}
   \caption{Performance at different ranks on the VIPeR dataset with different amount of training data. $p$ denotes the number of training samples and $r$ denotes the rank. It can be seen that the proposed method outperforms the best performing metric learning algorithm for person re-identification. }
\label{tab:numtrain}
\centering
\medskip\noindent
\begin{tabularx}{01\textwidth}{c *{16}{|Y}}
\toprule
Algorithm  & \multicolumn{4}{c|}{p=100} & \multicolumn{4}{|c|}{p=200} & \multicolumn{4}{|c|}{p=432}  & \multicolumn{4}{|c}{p=532}\\
		\cmidrule(lr){1-17} 					\cmidrule(l){6-9}			 \cmidrule(l){10-13} 		\cmidrule(l){14-17}
  &  r=1 &  r=5 &  r=10 &  r=20   &  r=1 & r=5 &  r=10 & r=20   &  r=1 & r=5 &  r=10 & r=20   &  r=1 & r=5 &  r=10 & r=20\\
\midrule
 SVMML \cite{ZhenliShiyu_CVPR2013}  & 9.7 &  30.1 &  43.6  & 60.0 		& 16.3 &  43.9  & 58.7  & 75.0 		&  33.5  & 68.9  & 84.0  & 92.8 	& 46.4  & 82.6 &  91.1  & 97.4 \\
 kLFDA \cite{eccv14prid}  			& 15.9 &  38.8  & 52.7 &  67.9	 	& 25.4 &  53.3  & 68.3  & 82.3 		& 45.2   & 81.2  & 91.4  & 97.2     & 56.4  & 90.8 & 96.6   & 99.3 \\ 
 MFA \cite{eccv14prid}  			& 15.5 &  38.3 &  52.0  & 67.7		&  25.0 &  52.7&   68.3&   82.3		&  45.0 &  80.6&   91.2&   97.2 	& 56.3 &  89.9 &  96.1 &  98.9\\
 {\bf Ours}   						& { \bf 17.7} &  {\bf43.2}  & {\bf58.1}  & {\bf70.7} 		&  {\bf27.5} &  {\bf59.3}  & {\bf72.9} &  {\bf86.3}		&  {\bf49.0} &  {\bf83.5}  & {\bf92.0} &  {\bf98.5} 	& {\bf62.0} & {\bf 91.6} &  {\bf 96.8} &  {\bf99.4}\\
\bottomrule
\end{tabularx}
\end{table*}

\section{Performance Analysis}
\label{sec:analysis}

%For a detailed evaluation of our approach, we conducted experiments with some baseline methods. First, we experimented on the proposed framework without using any kernel. Results indicate that  features demand a non-linear mapping for a better performance and learning invariant features in the kernel space is better than learning them in the original space. Another variant of the proposed work is the kLFDA framework extended to two layers. Compared to the performance achieved by the proposed framework, it can be seen that marginalization and invariance can perform better than the 2-layer kLFDA framework. This indicates that, advantages of training on large amount of data can be achieved by the proposed framework. It should also be noted that extending kLFDA to two layers does not give significant advantages over kLFDA directly applied to features.

For a detailed evaluation of our approach, we conducted experiments with some baseline methods. 

\subsection{Ours vs 2 layer kLFDA}
First, we extended the kLFDA framework in \cite{eccv14prid} to two layers. Compared to the performance achieved by the proposed framework, it can be seen that marginalization and invariance can perform better than the 2-layer kLFDA framework. This indicates that, advantages of training on large amount of data can be achieved by the proposed framework. It should also be noted that extending kLFDA to two layers does not give significant advantages over kLFDA directly applied to features. Individual local stripes from image pairs may not have enough discriminative capability compared to the whole image representation. Therefore, at the first stage, we extract only invariant features. Comparing the baselines marg\_kLFDA and 2 layer kLFDA, it can be inferred that extracting invariant information is more suitable at the local patch level.

\subsection{Ours without Invariance criterion (no\_Inv)}

The next baseline is no\_Inv where no invariance criterion is enforced. Intuitively, cross-view invariance is a key cue for person re-identification. This was validated from our experiments and results are reported in table \ref{actual_baseline} for all the datasets. While comparing to the final results obtained by the proposed framework, it can be seen that the invariance criterion improves the performance by around $2-4 \%$ for all the datasets.

\subsection{Ours without Marginalization (no\_Marg, marg\_SVMML and marg\_kLFDA)}

The next baseline we chose was the variant of the proposed approach without marginalization to show how much gain can be achieved by this technique. Since most of the person re-identification datasets are small, incorporating marginalization should give a better generalization capability. To analyze the advantages of marginalization, experiments were conducted in three stages. First, we conduct experiments on all datasets without marginalization on both layers. It can be seen from table \ref{actual_baseline} that, our approach substantially outperforms the baseline { no\_Marg}.

Further, we develop a variant without marginalization at the second layer, i.e. at the metric learning stage. Instead, the traditional kLFDA and SVMML were used for learning the metric. From table \ref{actual_baseline}, it can be observed that, performance of marg\_SVMML is close to the performance achieved by 2 layer kLFDA for iLIDS and CUHK01 datasets at all ranks but better for VIPeR and CAVIAR4REID. Even though, the SVMML performance is inferior to kLFDA (table \ref{table_Base}) when applied directly to the features, it can be well inferred from the results in table \ref{actual_baseline} that the proposed invariant feature learning framework with marginalization substantially helps the SVMML framework to perform better. But the performance of marg\_kLFDA is better than marg\_SVMML which is a combination of the traditional SVMML with the first layer of the proposed framework. It is also noteworthy that marg\_kLFDA can perform better than the 2 layer kLFDA.

\begin{figure*}
\begin{center}
%\fbox{\rule{0pt}{2in} \rule{1\linewidth}{0pt}}
 \includegraphics[width=0.75\linewidth ]
                   {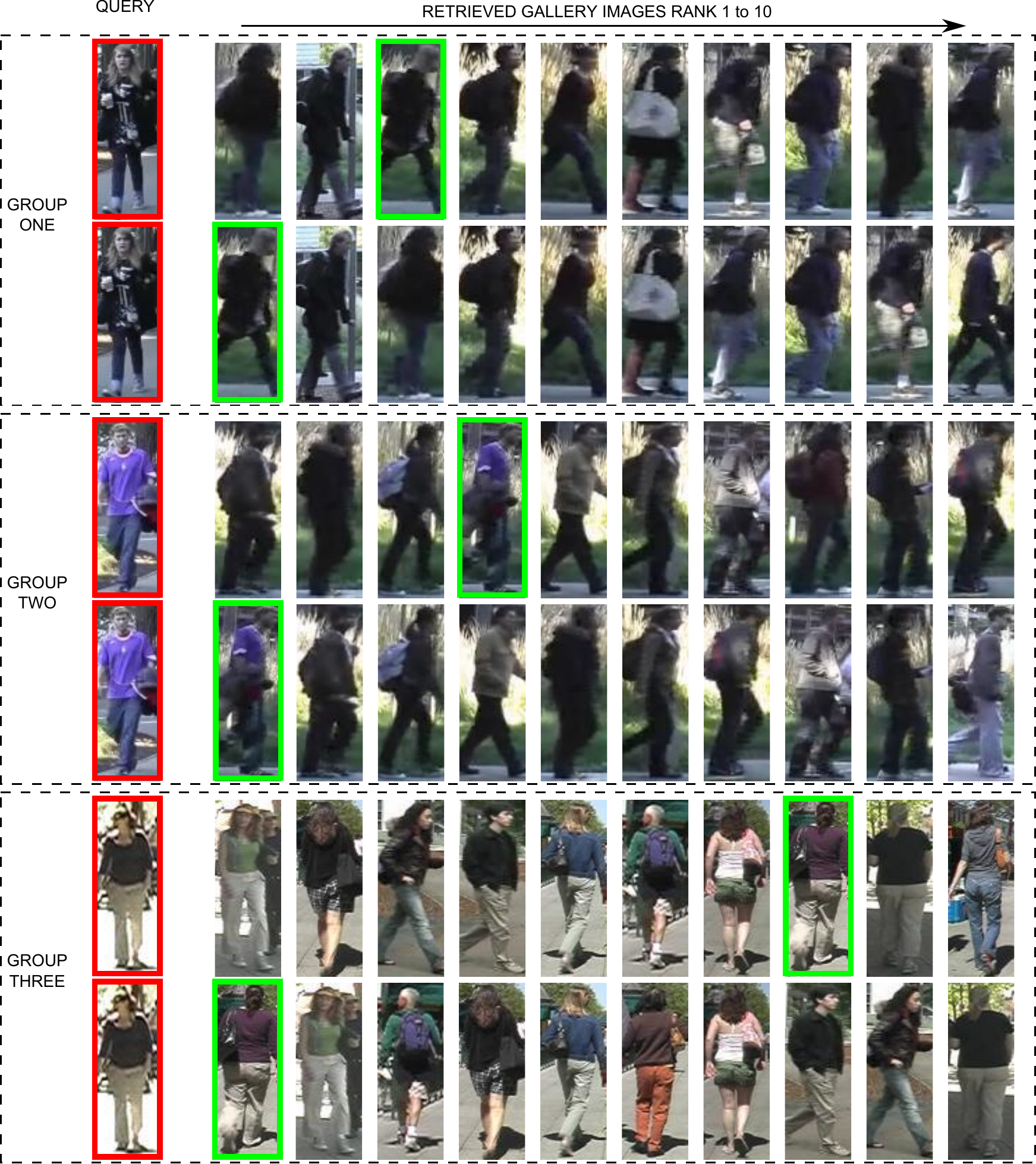}             
%                   {fig/framework/visualization.pdf}             
\end{center}
   \caption{Visualization of some results on SVMML and SVMML with marginalization. In each of the groups, as shown in the figure, the first row shows the top 10 retrieved matches for a query using marg\_SVMML. The second row shows the top 10 matches retrieved by the marginalized SVMML. It can be seen that several images within the top 5 ranks or top 10 ranks of marg\_SVMML were retrieved successfully at rank 1 by the proposed framework.}
\label{fig:visualization}
\end{figure*}

Finally, the proposed marginalized SVMML is used as the metric learning framework. From table \ref{actual_baseline}, it can be seen that the proposed metric learning framework outperforms marg\_SVMML in all scenarios indicating that marginalized SVMML can be advantageous. In most of the cases, marginalized SVMML outperforms marg\_kLFDA. For VIPeR and CUHK01 dataset, there is a notable improvement for the proposed framework over marg\_kLFDA, but for others, the improvement is not substantial. At some ranks marg\_kLFDA outperforms ours by a very small margin of $0.1 - 0.4 \%$. But in VIPeR and CUHK01, significant improvements can be seen at higher ranks. These results show that advantages of data augmentation can be achieved by the proposed framework based on marginalization and it achieves a better generalization over unseen data. Figure \ref{fig:visualization} shows a visualization of some of the results obtained by the proposed approach and marg\_SVMML. It can be seen that, many of the detections by marg\_SVMML in the top $5-10$ ranks were successfully detected at Rank 1 by our approach.

\subsection{Performance variation with number of training data on VIPeR}

More experiments were conducted on the VIPeR dataset using different number of training samples to validate the advantages of the proposed marginalized feature learning and metric learning framework. Table \ref{tab:numtrain} shows the results of the experiment and it can be seen that the proposed framework outperforms all the best performing metric learning methods for person re-identification. When $p=100$, as shown in the table \ref{tab:numtrain}, it can be seen that the proposed algorithm outperforms the traditional SVMML by a very large margin. Compared to MFA and kLFDA, eventhough the rank 1 performance is improved by only $2\%$, at higher ranks, it can be seen that the performance improvement is significant ($3-6\%$). Similarly, when $p=200$, the performance improvement at higher ranks is improved by around ($3-7\%$) at higher ranks.

\subsection{Comparison with state-of-the-art methods}

Comparison with state-of-the-art methods is given in table \ref{viper_state} - \ref{cuhk_state} for different datasets. It can be seen that, individually, the proposed method outperforms all the recent approaches for person re-identification such as \cite{ejazdeep2015,yangcolor2014,salientcolorECCV14} and \cite{eccv14prid}. For clear distinction between individual methods and ensemble methods, we have separated them by a horizontal split in tables \ref{viper_state}, \ref{ilids_state} and \ref{cuhk_state}. However, combination of several methods are becoming popular for person re-identification in the recent literatures \cite{zhao2014learning,ensemble2015}. It can be seen that, for VIPeR dataset, a combination of \cite{zhao2014learning} and \cite{ZhenliShiyu_CVPR2013} achieved $43.4\%$ at Rank 1 where as another ensemble approach, \cite{ensemble2015} achieved $45.9\%$ at Rank 1. A combination of ours with \cite{ZhenliShiyu_CVPR2013} achieved the new state-of-the-art for this dataset - $47.9 \%$ at Rank 1. But for other datasets, such as iLIDS and CUHK01, it can be seen that \cite{ensemble2015} outperforms the combination proposed by us at Rank 1 and all ranks respectively. In \cite{ensemble2015}, a learning algorithm is proposed to combine the matching scores obtained from different set of features for a metric learning so as to improve the lower rank performances. However, in our framework, matching scores obtained for same features using different metric learning framework are added together after rescaling them from $0-1$ for each query image. We would like to point out that the algorithm proposed in \cite{ensemble2015} is in fact complementary to ours in the sense that for fusing the scores obtained from different features (or different metric learning algorithms), similar learning approaches can be used. But since this is beyond the scope of the proposed algorithm, the obtained results with out any learning mechanism for fusion is reported.

%\begin{table}[!t]
%
%
%	\renewcommand{\arraystretch}{1.3}
%	\setlength{\tabcolsep}{2.5pt}
%	\caption{Performance Comparison of different metric learning algorithms and the variants of the proposed method for the VIPeR dataset. Proposed mDAEI outperforms all the metric learning algorithms as well as the variants of this approach.}
%	\label{viper_base}
%	\centering
%	[VIPeR]{
%	\begin{tabular}{|c|c|c|c|c|c|}
%		\hline
%		\bfseries Method & \bfseries Rank 1 & \bfseries Rank 5 & \bfseries Rank 10 & \bfseries Rank 20
%		\\\hline\hline
%		PCCA  & 19.6  & 51.5  & 68.2  & 82.9  \\
%		\hline
%		{\begin{tabular}{@{}c@{}}Ours with \\ Linear kernel\end{tabular}}  & 20.7  &  49.7  &  61.2  &  75.3 \\
%		\hline
%		{\begin{tabular}{@{}c@{}}no\_Marg\end{tabular}}  & 33.2  & 68.5  & 80.8  &  91.8 \\
%		\hline
%		{\bf \begin{tabular}{@{}c@{}} Proposed \\ approach\end{tabular}} & { \bf 36.4}   & { \bf 70.4}  & { \bf 83.6}  & { \bf 93.4} \\
%		\hline
%	\end{tabular}	}
%\end{table}

\begin{table}[!t]

	\renewcommand{\arraystretch}{1.3}
	\setlength{\tabcolsep}{2.5pt}
	\caption{Performance Comparison of state-of-the-art algorithms for the VIPeR dataset. Proposed approach when combined with \cite{ZhenliShiyu_CVPR2013} outperforms all the existing state-of-the-art methods for VIPeR dataset. The performance of the proposed algorithm individually is also comparable to the previous state-od-the-art methods at higher ranks. Results for \cite{ensemble2015} and \cite{ejazdeep2015} were taken from the CMC graphs in the respective literature. NA - Not Available in the literature}
	\label{viper_state}
	\centering
	{
	\begin{tabular}{|c|c|c|c|c|c|}
		\hline
		\bfseries Method & \bfseries Rank 1 & \bfseries Rank 5 & \bfseries Rank 10 & \bfseries Rank 20
		\\\hline\hline
		{\begin{tabular}{@{}c@{}}Kernel \\ Descriptors \cite{bo2010kernel}\end{tabular}}  & 18.1  & 44.0  & 59.8  & 77.5 \\
		\hline		
		LFDA \cite{pedagadi2013local} & 24.1 & 51.2 & 67.1 & 82.0 \\
		\hline
		SSCDL \cite{liu2014semi} & 25.6 & 53.7 & 68.1 & 83.6  \\
		\hline	
		PatMatch \cite{zhao2013unsupervised} & 26.9 & 47.5 & 62.3  & 75.6 \\
		\hline			
		{\begin{tabular}{@{}c@{}}Mid-level \cite{zhao2014learning}\end{tabular}}  & 29.1 & 52.3 & 65.9  & 79.9  \\
		\hline
		SVMML \cite{ZhenliShiyu_CVPR2013} & 29.4 & 63.3 & 76.3  & 88.1 \\
		\hline		
%		Deep Re-ID \cite{li2014deepreid}  & 11.07  & 33.23  & 48.42  &  60.44 \\
%		\hline
		VWCM \cite{zhang2014novel} & 30.7 & 63.0 & 76.0  & 88.6 \\
		\hline
		SalMatch \cite{zhao2013person} & 30.2 & 52.3 & 65.5  &  79.1 \\
		\hline
%		Deep ML \cite{yi2014deep} & 34.5 & 60.1 & 74.4 & 84.2 \\
%		\hline
		Deep ML \cite{yi2014deep} & 28.2 & 59.3 & 73.5 & 86.4 \\
		\hline
 		DL 2015 \cite{ejazdeep2015} 	& 34.8	& 63.7 & 75.8 & NA \\
 		\hline				
%		JDRML \cite{jdrml} & 34.7 & 65.4 & 78.6 & 89.6 \\
%		\hline	
		CMWCE \cite{yangcolor2014} & 37.6 & 68.1 & 81.3 & 90.2 \\
		\hline		
		{\begin{tabular}{@{}c@{}}Salient \\ Color names \cite{salientcolorECCV14} \end{tabular}} &  37.8 & 68.5 & 81.2 & 90.4 \\
		\hline	
			{\bf \begin{tabular}{@{}c@{}} Ours \end{tabular}} & {\bf 39.3}   & {\bf  73.0}  & {\bf  84.6}  & {\bf 92.5} \\
			\hline \hline 
		{\begin{tabular}{@{}c@{}}Mid-level \cite{zhao2014learning} \\+ SVMML \cite{ZhenliShiyu_CVPR2013} \end{tabular}}  &  43.4 & 73.0 & 84.9  & 93.7  \\
		\hline						
%		{\begin{tabular}{@{}c@{}}Structured \\ Prediction \cite{zhang2014structured} \end{tabular}} & { \bf 38.9} & 67.4 & 80.4 & 90.2  \\
%		\hline
		{\begin{tabular}{@{}c@{}} { Metric }\\Ensembles\cite{ensemble2015}\end{tabular}} & {45.9 }  & {  77.5}  & {  88.9}  & { 95.8} \\
		\hline					
%		{\bf \begin{tabular}{@{}c@{}} margSingle-kLFDA\end{tabular}} & { 36.4}   & { 70.4}  & { 83.6}  & {  93.4} \\
%		\hline
		{\begin{tabular}{@{}c@{}} {marg\_kLFDA}\\+ SVMML\cite{ZhenliShiyu_CVPR2013}\end{tabular}} & {46.5 }  & {  74.1}  & { 86.4}  & { 95.1} \\
		\hline		
		{\begin{tabular}{@{}c@{}} {\bf Ours }+ SVMML\cite{ZhenliShiyu_CVPR2013}\end{tabular}} & {\bf 47.9 }  & { \bf 79.7}  & { \bf 90.2}  & { \bf 95.9}\\
		\hline	
%		{\begin{tabular}{@{}c@{}} {\bf Ours* }+ SVMML\cite{ZhenliShiyu_CVPR2013}\end{tabular}} & {\bf48.4 }  & { \bf 79.7}  & { \bf 90.2}  & { \bf 95.9} \\
%		\hline						
	\end{tabular}	}
\end{table}
\begin{table}[!th]
	\renewcommand{\arraystretch}{1.3}
	\setlength{\tabcolsep}{2.5pt}
	\caption{Performance Comparison of state-of-the-art algorithms for the CAVIAR4REID dataset in the single-shot setting. The proposed approach outperforms all the methods except MFA at rank 1. A combination of our approach with kLFDA \cite{eccv14prid} achieves the state-of-the-art results for CAVIAR4REID dataset.}
	\label{caviar_state}
	\centering	
	{
	\begin{tabular}{|c|c|c|c|c|}
			\hline
			\bfseries Method & \bfseries Rank 1 & \bfseries Rank 5 & \bfseries Rank 10 & \bfseries Rank 20
			\\\hline\hline
			LFDA \cite{pedagadi2013local} & 32.0  & 56.3  & 70.7  & 87.4  \\
			\hline
			kLFDA \cite{eccv14prid} & 35.9  & 63.6  & 77.9  & 91.2 \\
			\hline			
			SVMML \cite{ZhenliShiyu_CVPR2013} & 31.2  & 62.8  & 78.5  &  94.2  \\
			\hline
			MFA \cite{eccv14prid} & {  40.2}  & 70.2  & 83.9  & 95.1  \\
			\hline
		{\bf \begin{tabular}{@{}c@{}} Ours \end{tabular}} & { 39.9}   & { 73.2}  & {  88.4}  & { 95.7} \\
		\hline	
		{\bf \begin{tabular}{@{}c@{}} Ours + kLFDA \cite{eccv14prid} \end{tabular}} & { \bf 45.1  }   & { \bf 76.8}  & { \bf 88.9}  & { \bf 97.4} \\
		\hline		
		
		\end{tabular}
	}	
\end{table}
\begin{table}[!th]
	\renewcommand{\arraystretch}{1.3}
	\setlength{\tabcolsep}{2.5pt}
	\caption{Performance Comparison of state-of-the-art algorithms for the iLIDS dataset in the single-shot setting. Proposed approach performs comparably to \cite{ensemble2015}. The combination of ours with kLFDA \cite{eccv14prid} achieves state-of-the-art results at higher ranks but performs inferior to Metric Ensembles at rank 1. We believe that this is due to the learned score combining mechanism proposed in \cite{ensemble2015}.}
	\label{ilids_state}
	\centering	
%	{
	\begin{tabular}{|c|c|c|c|c|}
			\hline
			\bfseries Method & \bfseries Rank 1 & \bfseries Rank 5 & \bfseries Rank 10 & \bfseries Rank 20
			\\\hline\hline
			PRDC \cite{prdczhengpami2012} & 37.8  & 63.7  & 75.1  & 88.4  \\
			\hline
			kLFDA \cite{eccv14prid} & 38.0  & 65.1  & 77.4  &  89.2  \\
			\hline
		{\bf \begin{tabular}{@{}c@{}} Ours\end{tabular}} & {\bf  39.5}   & {\bf 70.4}  & {\bf 81.0}  & { \bf 91.4} \\
		\hline \hline			
			Metric Ensembles \cite{ensemble2015} & { \bf 50.3}  & 71.9  & 80.6  & 91.3  \\
			\hline
		{\bf \begin{tabular}{@{}c@{}} Ours + kLFDA \cite{eccv14prid}\end{tabular}} & {  47.7}   & { \bf 73.1}  & { \bf 84.9}  & { \bf 93.9} \\
		\hline		
		\end{tabular}
%	}	
\end{table}
  \begin{table}[!th]
  	\renewcommand{\arraystretch}{1.3}
  	\setlength{\tabcolsep}{2.5pt}
  	\caption{Performance Comparison of state-of-the-art algorithms for the CUHK01 dataset in the single-shot setting. The proposed approach outperforms all the methods at all ranks. Results for \cite{ensemble2015} and \cite{ejazdeep2015} were taken from the CMC graphs in the respective literature.}
  	\label{cuhk_state}
  	\centering	
  %	{
  	\begin{tabular}{|c|c|c|c|c|}
  			\hline
  			\bfseries Method & \bfseries Rank 1 & \bfseries Rank 5 & \bfseries Rank 10 & \bfseries Rank 20
  			\\\hline\hline
  			SDALF \cite{sdalf} & 9.9  & 22.6  & 30.3  & 41.0  \\
  			\hline
  			eSDC \cite{zhao2013unsupervised} & 19.7  & 32.7 & 40.3  & 50.6  \\
  			\hline
  			LMNN \cite{Weinberger2009LMNN} & 13.5  & 31.3  & 42.3  & 54.1  \\
  			\hline			
  			ITML \cite{itml} & 16.0  & 35.2  & 45.6  & 59.8  \\
  			\hline
  			SalMatch \cite{zhao2013person} & 28.5  & 45.9  & 55.7  & 68.0  \\
  			\hline
  			Midlevel \cite{zhao2014learning} & {\bf 34.3}  & {\bf 55.1}  & 65.0  & 74.9  \\
  			\hline
		{\bf \begin{tabular}{@{}c@{}} Ours\end{tabular}} & {29.2 }& {54.7 }& {\bf 66.3}   & {\bf 77.6} \\
		\hline	\hline  			
%  			visWord	 \cite{zhang2014novel}						& 44.03 & 70.7	& 79.2 	& NA \\
%  			\hline
%  			DL 2015 \cite{ejazdeep2015} 	& 47.5	& 72.0 & 80.5 & NA \\
%  			\hline
  			Metric Ensembles \cite{ensemble2015} & {\bf 53.4}  &{\bf  76.7 } &{\bf  84.4 } &{\bf  90.1 } \\
  			\hline	
  																	
  		{\bf \begin{tabular}{@{}c@{}} Ours + kLFDA \cite{eccv14prid}\end{tabular}} & 39.5 &  62.1  & 74.3 &  82.9 \\
  		\hline
  		\end{tabular}
  %	}	
  \end{table}

\section{Conclusion}
\label{sec:concl}

We proposed a novel invariant feature learning framework with marginalization for person re-identification. 
To handle the non-linearities introduced by variations in pose, illumination and environment,  
Local features extracted from the images and are first transformed into a kernel space. A linear transformation is learned in this kernel space to capture invariant information by using labeled image pairs. Since the amount of labeled pairs is less, we propose a novel objective function with marginalization to reap the benefits of training on {\it infinite} data. These mapped local features are concatenated to form the whole image representation and fed into a metric learning framework for classification. To achieve better generalization over the test set, we proposed a marginalized metric learning framework based on the popular SVM Metric Learning. The proposed approach was tested on four challenging publicly available datasets and our experiments show that learning invariant representations from the labeled images can improve the person re-identification performance. Additionally, the marginalization technique is shown to be advantageous when there is a lack of training data. Our comparison with state-of-the-art algorithms show that our method has a lot of future prospects.

%\vspace{-15pt}
%{\small
%\bibliographystyle{ieee}
%\bibliography{egbib}
%}

%\end{document}

\ifCLASSOPTIONcaptionsoff
  \newpage
\fi

% references
\bibliographystyle{IEEEbib}
\bibliography{references}

%\begin{IEEEbiography}{}
%\end{IEEEbiography}

\end{document}